\documentclass{article}

\PassOptionsToPackage{numbers, sort, compress}{natbib}
\usepackage[preprint]{neurips_2026}

\usepackage[utf8]{inputenc} %
\usepackage[T1]{fontenc}    %
\usepackage{hyperref}       %
\usepackage{url}            %
\usepackage{booktabs}       %
\usepackage{amsfonts}       %
\usepackage{nicefrac}       %
\usepackage{microtype}      %
\usepackage{xcolor}         %
\usepackage{xspace}
\usepackage{graphicx}
\usepackage{amsmath}
\usepackage{cleveref}   %
\usepackage{soul}
\usepackage{amssymb}
\usepackage{multirow}
\usepackage[normalem]{ulem}

\title{$\phi$-Noise: Training-Free Temporal Video Conditioning via Phase-Based Noise Manipulation }

\author{%
  Ofir Abramovich\footnotemark[1] \\
  Canvas-Lab \\
  Department of Computer Science \\
  Reichman University \\
  \And
  Nadav Z.~Cohen\thanks{Denotes Equal Contribution and fondness of cats} \\
  Canvas-Lab\\
  Department of Computer Science\\
  Reichman University\\
  \And
Adi Rosenthal\footnotemark[1] \\
  Canvas-Lab \\
  Department of Computer Science \\
  Reichman University \\
  \And
  Ariel Shamir \\
  Canvas-Lab \\ 
  Department of Computer Science \\
  Reichman University \\
}

\begin{document}

\maketitle

\begin{abstract}

Latent video diffusion models generate videos by progressively transforming Gaussian noise into realistic samples conditioned on text or visual inputs. However, existing conditioning methods often require additional training and computational overhead.
Motivated by recent findings on the importance of frequency components in generative models,
we propose a simple, training-free approach for motion-conditioned video generation by injecting low-frequency phase information from a reference video directly into the diffusion noise latents. Our method transfers motion cues without modifying the model architecture or inference pipeline.
Using several applications, we demonstrate effective control over both appearance and dynamics in generated videos, while achieving competitive or superior results compared to more complex conditioning approaches.
\end{abstract}

\section{Introduction}
Latent diffusion models have become the dominant paradigm for visual content generation, achieving remarkable success in image synthesis~\cite{Rombach_2022_CVPR, podell2023sdxlimprovinglatentdiffusion, flux2024, wu2025qwen, sd35, stable_cascade}. More recently, diffusion-based video generation models extended these capabilities to temporally coherent multi-frame synthesis~\cite{ltx, wan2025wanopenadvancedlargescale, yang2025cogvideoxtexttovideodiffusionmodels, lumier, ho2022imagenvideohighdefinition, blattmann2023stablevideodiffusionscaling}. These models progressively transform white Gaussian noise in latent space into structured visual outputs conditioned on textual prompts.

While text prompts provide high-level semantic guidance, recent works explored additional conditioning mechanisms for more precise control. In the image domain, methods for structure and style conditioning enable control beyond natural language descriptions~\cite{control_net, ip_adapter, t2i_adapter, stylealigned, blora, conditional_balance}. Extending such control to video generation is substantially more challenging, as videos require modeling both spatial appearance and temporal dynamics, including motion and camera behavior.

To address this, prior works introduced methods for conditioning either spatial~\cite{chen2023controlavideo, videocontrolnet, animate_anyone} or temporal~\cite{motionprompting, motion_inversion, pondaven2025ditflow} aspects of generated videos. However, many approaches rely on specialized architectures, additional training, or computationally expensive inference-time operations, increasing both complexity and runtime.

In this work, we introduce $\phi$-Noise, a training-free video conditioning framework that manipulates the input noise prior to diffusion using phase information ($\phi$) extracted from a reference video, without introducing significant runtime or memory overhead. While diffusion noise is typically treated as a purely stochastic source, we show that its low-frequency components strongly influence global spatial and temporal structure. Motivated by this observation, we selectively modify the low-frequency phase of the input noise to inject structural and temporal biases into the generation process without changing the model architecture or inference pipeline.

Despite its simplicity, our method enables effective motion and spatial conditioning without additional training or expensive signal analysis. We demonstrate applications including motion conditioning, spatial conditioning, and cut-and-drag generation, achieving competitive or superior results compared to recent approaches. We further show that manipulating different frequency bands provides controllable variations in the conditioning behavior.

Our contributions are summarized as follows:
\begin{itemize}
\item We analyze diffusion video generation from a frequency-domain perspective, studying how temporal frequency manipulation shapes generated motion and affects latent energy evolution throughout the diffusion process. 

\item We propose $\phi$-Noise, a training-free and low-overhead framework for spatial and temporal video conditioning through simple noise-phase manipulation.
\item We demonstrate that our approach generalizes across tasks and architectures, and can also be applied to conditional image generation models.
\end{itemize}

Code and implementation details are available on our project page.

\begin{figure*}[t]
    \centering
    \includegraphics[width=\textwidth]{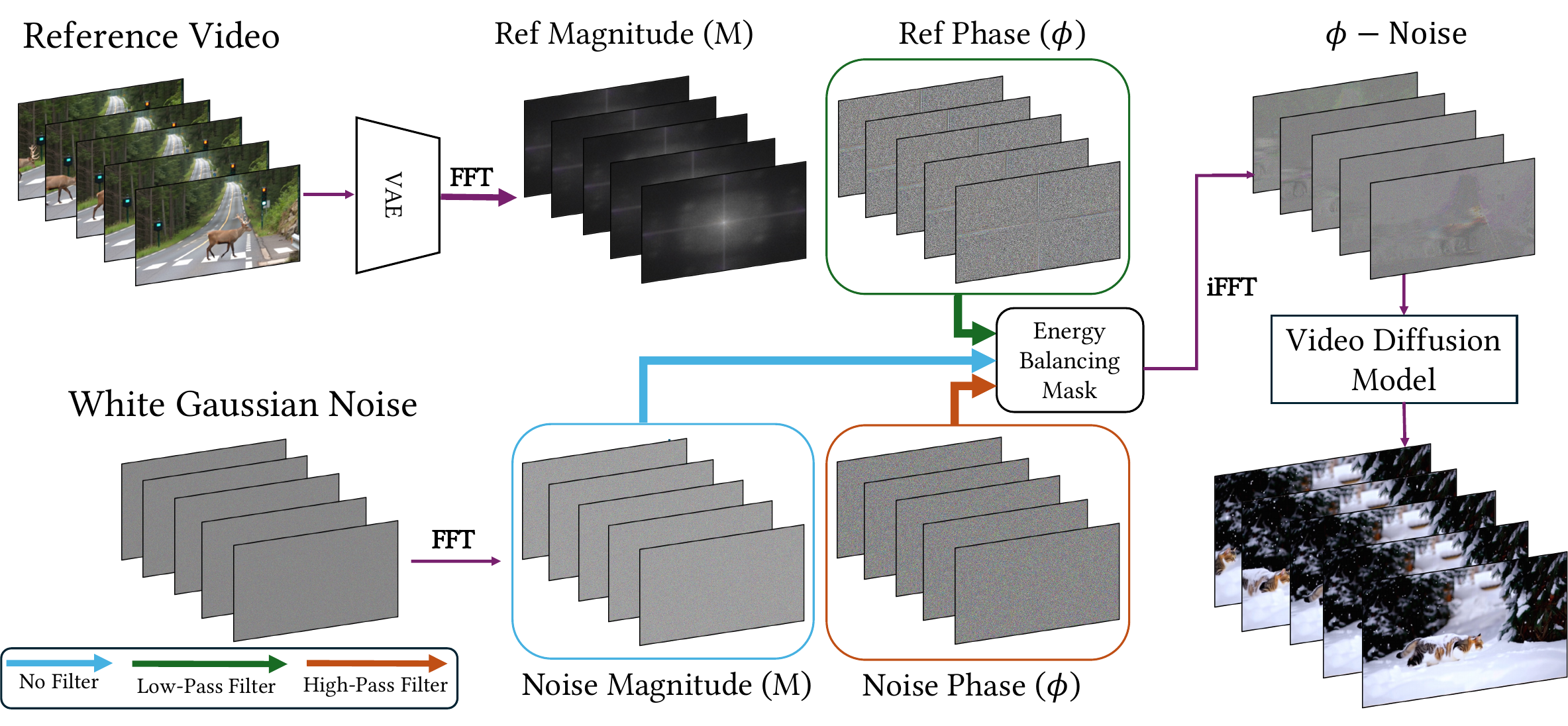}
    \caption{\textbf{Method Overview.} We calculate the frequency decomposition of both noise an signal using the Discrete-Fourier Transform to phase and magnitude. Then, we replace the low-frequencies phase information of the noise with those of the conditional input, and normalize total energy of the reconstructed noise. The output noise is then used as input to the generation model. Note that we show the frames of the original video for visualization but in practice work in latent space. }
    \label{fig:tasks}
\end{figure*}

\section{Related Work}
\label{related_work}

\subsection{Diffusion-based Video Generation}

\vspace{-0.25cm}
Following the success of latent image diffusion models~\citep{Rombach_2022_CVPR, podell2023sdxlimprovinglatentdiffusion, stable_cascade, flux2024, wu2025qwen, sd35}, video generation has naturally emerged as an extension to the temporal domain, aiming to synthesize videos by progressively denoising spatio-temporal white Gaussian noise. Early works explored unconditional or weakly conditioned video diffusion models~\cite{ho2022videodiffusion, he2023latentvideodiffusionmodels}, demonstrating promising visual quality but limited controllability.

Subsequent approaches leveraged pretrained text-to-image models to enable controllable video synthesis. These include methods that inject temporal information into independently generated frames~\cite{text2vide_zero}, introduce trained modules that enforce temporal consistency~\cite{align_your_latents, make_a_video, ho2022imagenvideohighdefinition}, or combine image-based backbones with video-specific architectures~\cite{tune_a_video} to benefit from strong spatial priors learned on large-scale image datasets, enabling high-quality generation with limited video data.

More recent works further improve fidelity, scalability, and motion coherence~\cite{lumier, guo2024animatediff, yang2025cogvideoxtexttovideodiffusionmodels, chen2024videocrafter2, openai_sora_2024}, while still relying primarily on text conditioning or hybrid image-video training. However, text alone remains insufficient for specifying fine-grained spatial structure and complex temporal dynamics such as object motion and camera trajectories.

\subsection{Visual Conditioning in Video Generation}

\vspace{-0.25cm}
To address the limitations of text conditioning, visual conditioning has emerged as a more expressive mechanism for controlling generated videos. Existing approaches can be broadly divided into \textit{structural} and \textit{temporal} conditioning, which are typically addressed separately and often require task-specific designs.

\textbf{Structural conditioning.}
Building on image-based conditioning methods~\cite{control_net, ip_adapter, t2i_adapter}, prior works incorporate visual signals such as edges, depth, or pose into video generation. These signals are typically injected via auxiliary networks trained to guide the generation process~\cite{chen2023controlavideo, animate_anyone}, or applied frame-wise using pretrained modules such as ControlNet~\cite{control_net}~\cite{videocontrolnet}. Other approaches adapt pretrained image diffusion models to video generation~\cite{tune_a_video, text2vide_zero}, reusing image conditioning mechanisms.

More recent methods directly condition on images or videos~\cite{tokenflow2023, wan2025wanopenadvancedlargescale, ltx, blattmann2023stablevideodiffusionscaling}, using them as structural references. Additional works explore interactive or spatial control signals, such as drag-based editing or region-level guidance~\cite{shi2023dragdiffusion, mou2023dragondiffusion, diffuhaul}, extending structural conditioning to more flexible user control.

\textbf{Temporal conditioning.}
Controlling motion dynamics remains more challenging and is often addressed separately from structure. Early methods rely on explicit motion representations such as optical-flow or camera trajectories~\cite{motionprompting, only_flow, direct_a_video}, which require additional estimation or user specification. More recent approaches instead learn implicit motion representations through latent alignment~\cite{moft}, inversion~\cite{motion_inversion, ling2024motionclone}, or motion descriptor optimization~\cite{space_time}. Other works incorporate camera control or trajectory-guided generation~\cite{he2024cameractrl, deng2023dragvideo}, and optimization based tempo control~\cite{schiber2026tempocontroltemporalattentionguidance}, further highlighting the diversity of task-specific solutions.

Despite their effectiveness, most existing approaches focus on either structural or temporal control and rely on additional components such as auxiliary networks, inversion procedures, attention manipulation, or specialized representations. This often increases computational overhead and limits generality across tasks and architectures.

\subsection{Latent Noise Manipulation}

Diffusion models generate visual outputs by progressively transforming white Gaussian noise into structured samples. Although the input noise is typically treated as unstructured randomness, several works have explored manipulating it to guide the generation process.

FreeInit~\cite{wu2024freeinitbridginginitializationgap} injects low-frequency components from inverted videos into the noise to improve temporal coherence, while FreqPrior~\cite{yuan2025freqprior} decomposes noise into frequency bands to enhance detail generation. Other approaches employ noise warping~\cite{noise_warping, how_i_warped, go_with_the_flow} for temporal consistency and motion control, and Time-to-Move~\cite{time2move} utilizes SDEdit~\cite{sdedit} for localized motion editing. In the image domain, noise manipulation has also been used for image editing~\cite{fds} and conditioning on hand-drawn color maps~\cite{colorful-noise}. Additional works further challenge the assumption of white Gaussian noise by exploring alternative noise distributions for generative models~\cite{voleti2022scorebaseddenoisingdiffusionnonisotropic, rissanen2023generativemodellinginverseheat, blue_noise}.

Most closely related to our work, NeuralRemaster~\cite{neuralremaster} manipulates Fourier components of the input noise to inject spatial structure via phase information. Similar to this line of work, we leverage phase information for conditioning; however, unlike NeuralRemaster, our method is entirely training-free and operates solely by modifying the input noise prior to generation.

\textbf{Our perspective.}
Building on these observations, we directly manipulate the frequency decomposition of the input noise by injecting low-frequency components from conditioning signals prior to diffusion. Since low frequencies capture coarse spatial structure and dominant temporal dynamics, this enables control over both appearance and motion during generation.

Unlike prior approaches, our method provides a unified conditioning framework operating over the spatial and temporal dimensions of the noise. It is entirely training-free, introduces minimal computational overhead, and remains agnostic to the underlying model architecture.

\section{Analysis}
\label{sec:analysis}

In this section, we analyze the spectral behavior of white Gaussian noise in a Video Diffusion model and present experiments to build intuition for the proposed method. We define frequency extraction operators in the context of video latents, followed by an in-depth investigation into phase-based noise manipulation. We specifically examine how this manipulation affects the latent phase distribution and the critical importance of maintaining spectral energy balance to prevent generative divergence.

\vspace{-0.25cm}
\subsection{Preliminaries}
\vspace{-0.25cm}
Our primary goal is motion dynamic transfer. Hence, to isolate motion dynamics as much as possible, we utilize in our analysis a neutral-background reference video featuring simple object movements (e.g., \textit{``A ball bouncing up and down''}). We examine how temporal spectral manipulation can transfer these motion patterns to a newly generated video.

Let $V \in \mathbb{R}^{T \times W \times H \times C}$ be a sequence of $T$ frames. A latent diffusion model encodes $V$ into a latent tensor $\mathbf{v} \in \mathbb{R}^{t \times w \times h \times d}$. We apply the Discrete Fourier Transform (DFT) to map this latent signal to the frequency domain along either the 1D \textit{temporal} ($t$) or 2D \textit{spatial} ($w, h$) dimensions, denoted as $\mathcal{F}_T$ and $\mathcal{F}_S$, respectively. We first analyze temporal decomposition for motion transfer, then extend the approach to the spatial domain for structural conditioning.

Given initial white Gaussian noise $\mathbf{z}$ of the same dimensionality as $\mathbf{v}$, we define their frequency representations $\mathbf{\tilde{v}}, \mathbf{\tilde{z}}$ by applying the DFT along the temporal dimension:
\begin{equation}
    \left\{\mathbf{\tilde{v}}, \mathbf{\tilde{z}}\right\} = \mathcal{F}_{T}\left(\left\{{\mathbf{v}, \mathbf{z}}\right\}\right).
\end{equation}
The complex-valued representations are then decomposed into magnitudes $M^\mathbf{v}, M^\mathbf{z}$ and phases $\phi^\mathbf{v}, \phi^\mathbf{z}$.
Following Parseval’s Theorem~\cite{https://doi.org/10.1112/plms/s2-25.1.237}, the total energy of a discrete signal is preserved (up to a scaling constant) between the spatial and frequency domains. We define the energy $\text{E}$ of the latent $\mathbf{z}$ as the sum of its squared components:
\begin{equation}
    \text{E}(\mathbf{z}) = \sum_{i} \mathbf{z}_i^2 \propto \text{E}    (\mathbf{\tilde{z}}) = \sum_{i} |\mathbf{\tilde{z}}_i|^2.
\end{equation}
In this context, $\text{E}(\mathbf{z})$ represents the total ``strength'' of the signal.

\vspace{-0.25cm}
\subsection{Phase Manipulation of White Gaussian Noise}
\label{subsec:phase}

\vspace{-0.25cm}
In signal processing, the phase spectrum typically captures structural information, whereas the magnitude spectrum captures the distribution of energy across scales. In the temporal domain, this structure corresponds directly to motion patterns. Our hypothesis is that by substituting the low-frequency phase of $\mathbf{z}$ with the phase of a reference video $\mathbf{v}$ we can control the motion dynamics of the video generated by $\mathbf{z}$. We focus on low frequencies, as they represent the global, coarse motion trajectories most critical for temporal coherence.

Let $k$ be the frequency cutoff point ($0 < k \le t$). We define $\mathbf{\tilde{z}}_{k}$ as the latent resulting from substituting the $k$ lowest temporal frequency phases of $\mathbf{z}$ with those of $\mathbf{v}$:
\begin{equation}
    \mathbf{\tilde{z}}_{k} = M^\mathbf{\tilde{z}} \odot e^{i\phi^{\mathbf{\tilde{z}}_k}},\ \text{where}\     \phi^{{\mathbf{\tilde{z}}}_k}_f =
    \begin{cases}
        \phi^\mathbf{v}_f & \text{if } f \le k \\
        \phi^\mathbf{z}_f & \text{if } f > k
    \end{cases},
    \label{eq:phase_subs}
\end{equation}
for each frequency index $f$.

\begin{figure*}[t]
    \centering
    \includegraphics[width=\textwidth]{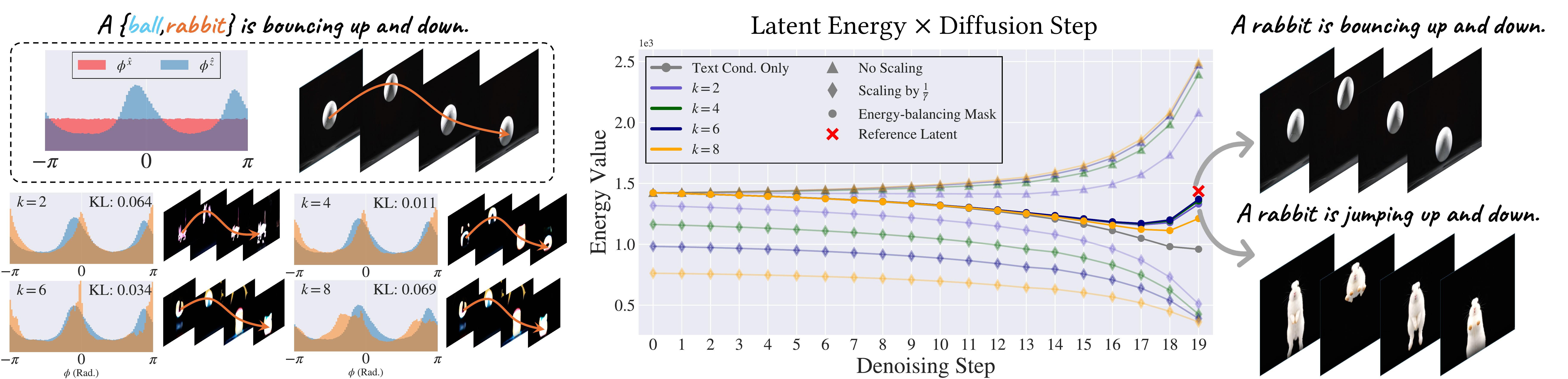}
    \caption{\textbf{Phase and Energy Analysis.} We analyze the impact of substituting $k$ low-frequency phase components in the latent space prior to denoising. (Left) Comparison of phase distributions between the reference video ({\color{blue}blue}) and the generated outputs ({\color{orange}orange}). (Middle) Evolution of latent energy across denoising timesteps for various $k$ values (colors) and scaling settings (markers). The red symbol (${\color{red} \times}$) denotes the reference energy $\text{E}(\mathbf{x})$. (Right) Qualitative Comparison. Applying our energy-balancing mask $\Phi$ preserves signal energy, ensuring stable denoising and high-fidelity motion transfer that faithfully follows the reference dynamics. We recommend zooming-in for a better view.}    \label{fig:phase_analysis}
    
    \vspace{1em} %
    
    \includegraphics[width=\textwidth]{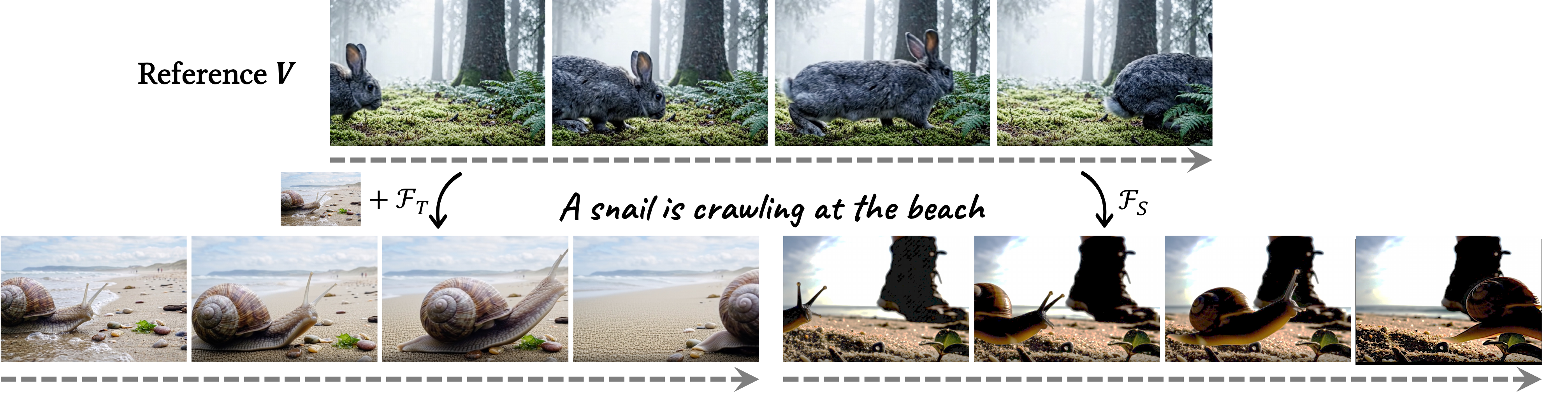}
    \caption{\textbf{Global Structure Transfer.} In addition to motion, we propose two methods for global structure transfer: (1) Image-to-Video (I2V) Motion Transfer, by utilizing an input image to fully preserve scene characteristics, layout, and identities (left); and (2) Implicit Temporal Conditioning, where the spatial layout and dynamics are preserved from the reference video $V$ (right).}
    \label{fig:spatial_comp}
\end{figure*}

Next, we use $\mathbf{\tilde{z}}_k$ to generate a video conditioned on a text prompt. This operation is non-trivial, as video diffusion models are trained on i.i.d.\ white Gaussian noise with uniformly distributed phase, $\phi \sim \mathcal{U}(-\pi, \pi)$. In \cref{fig:phase_analysis} (left), we analyze the phase distributions by measuring the Kullback–Leibler (KL) divergence between generated outputs and reference videos. While the injected phase successfully transfers motion (orange trajectories), the resulting videos exhibit severe saturation and poor prompt alignment (\cref{fig:phase_analysis}, bottom-left grid). As shown in the energy evolution plot (\cref{fig:phase_analysis}-right, $\blacktriangle$-shaped plots), injecting low frequencies alters the latent energy throughout denoising, with larger $k$ values causing significant energy divergence and out-of-distribution artifacts.

\vspace{-0.25cm}
\subsection{Energy Effect of Spectral Manipulation}
\label{subsec:a:energy}

\vspace{-0.25cm}
The divergence observed in the previous section suggests that phase substitution disrupts the expected energy profile of the noise. To investigate this, we first attempted to scale the manipulated low-frequency magnitudes by a constant factor $1/\gamma$. However, as shown in Fig. \ref{fig:phase_analysis} (middle, $\blacklozenge$-shaped plots), this leads to an energy collapse--dropping to ~35\% of the reference level-- which leads to a collapse, where no meaningful video is produced.

To stabilize the denoising process, we propose a Spectral-Temporal Energy Balancing Mask $\Phi \in \mathbb{R}^t$, which ensures that while we scale down the high-energy low frequencies (by $1/\gamma$), we compensate by scaling the remaining frequencies by a factor $\beta$ to preserve the total energy $\text{E}(\mathbf{z})$.

\begin{equation}
    \Phi(\mathbf{\tilde{z}},k,\gamma)_f =
    \begin{cases}
        1/\gamma & \text{if } f \leq k \\
        \beta & \text{if } f > k
    \end{cases}, \quad \text{where }
        \beta = \sqrt{ \frac{\text{E}(\mathbf{\tilde{z}}) - \frac{\text{E}(\mathbf{\tilde{z}}_{\text{l}})}{\gamma^{2}}}{\text{E}(\mathbf{\tilde{z}}_{\text{h}})} },
    \label{eq:eb_mask}
\end{equation}
where $f$ is the frequency index, and $\mathbf{\tilde{z}}_{\text{l}},\mathbf{\tilde{z}}_{\text{h}}$ are the $k$ and $(t-k)$ lowest and highest frequencies of $\mathbf{z}$, respectively. For a detailed derivation of $\beta$, please refer to the Appendix A.

By construction of this mask, we ensure that energy is preserved:
\begin{equation}
    \text{E}(\mathbf{\tilde{z}}_{k} \odot \Phi) = \text{E}(\mathbf{\tilde{z}}).
\end{equation}
Applying $\Phi$ effectively ``re-whitens'' the noise energy across the spectrum. As shown in \cref{fig:phase_analysis} (middle), this stabilization keeps the denoising process in-distribution, yielding high-quality videos that follow the reference motion without the artifacts of raw phase substitution.

\vspace{-0.25cm}
\subsection{Structure Conditioning}
\label{subsec:exp_structure}

\vspace{-0.25cm}
In complex real-world videos, where the background contains intricate details, temporal frequencies alone may be insufficient for structure transfer and generation. The temporal domain cannot explicitly define the underlying spatial geometry. To address this, as will be shown in \cref{method}, we must incorporate an additional input structural condition, such as a reference image, to provide explicit spatial scene information. This serves as a ``structured guideline'' to anchor geometric layout and identity details, as demonstrated In~\cref{fig:spatial_comp} (bottom left).

Alternatively, we apply a 2D spatial DFT, $\mathcal{F}_S$, to map each frame's spatial domain to the frequency domain. This provides stronger structural conditioning per frame, which implicitly preserves video motion through continuous alignment. By substituting the $k$ lowest spatial phase frequencies via a radial mask, we capture the global layout while allowing textures to adapt to the target prompt, as shown in \cref{fig:spatial_comp} (bottom right) (e.g., grass $\rightarrow$ sand; tree $\rightarrow$ shoe).

\section{Method}
\label{method}

Building on our findings in \cref{sec:analysis}, we formulate a general and efficient framework for manipulating specific noise latent frequencies based on a conditioning reference video. We propose two methods that leverage the frequency domain. One applies to the temporal dimension, and the other to the spatial dimensions to anchor both motion and structure, while regulating the signal's energy to ensure sampling stability and robust generation.
\vspace{-0.25cm}

\subsection{Spectral Decomposition}
\vspace{-0.25cm}
Given a reference video latent $\mathbf{v}$ and a gaussian noise latent $\mathbf{z}$, we first transform both into the frequency domain using a temporal or spatial DFT-- $\mathcal{F}_{D\in\left\{\text{T,S}\right\}}$:
\begin{equation}
    \mathbf{\tilde{v}} = \mathcal{F}_{\text{D}}(\mathbf{v}), \quad \mathbf{\tilde{z}} = \mathcal{F}_{\text{D}}(\mathbf{z}).
\end{equation}
Next, we decompose these spectral coefficients to magnitude $M$ and phase $\phi$:
\begin{equation}
    \mathbf{\tilde{v}} = M^\mathbf{v} \odot e^{i\phi^\mathbf{v}}, \quad \mathbf{\tilde{z}} = M^\mathbf{z} \odot e^{i\phi^\mathbf{z}}.
\end{equation}

\subsection{Phase Substitution and Energy Balancing}
\vspace{-0.25cm}
To transfer the structural motion of the reference to the noise latent, we substitute the phase of $\mathbf{\tilde{z}}$ with that of $\mathbf{\tilde{v}}$ up to a frequency cutoff $k$, following~\cref{eq:phase_subs}, to produce $\mathbf{\tilde{z}}_k$.

As shown in \cref{subsec:a:energy}, phase manipulation can disrupt the signal’s energy evolution throughout the generation process. To preserve stability, we compute the spectral energy balancing mask $\Phi(\mathbf{\tilde{z}}, \gamma, k)$ (\cref{eq:eb_mask}) and apply it to $\mathbf{\tilde{z}}_k$ to ensure energy conservation. Finally, the latent is mapped back to the spatial domain using the Inverse DFT:
\begin{equation}
    \mathbf{z}^\Phi = \mathcal{F}^{-1}_{\text{D}}\left(\Phi \odot \mathbf{\tilde{z}}_k\right).
    \label{eq:energy_conservation_mask}
\end{equation}
The resulting latent $\mathbf{z}^\Phi$ serves as the initialization for the denoising process, biasing the generation toward the motion of the reference video while remaining within the model's learned distribution. Since applying the DFT and its inverse is computationally negligible compared to even a single diffusion iteration, the proposed method introduces negligible runtime and memory overhead, as it does not intervene in the diffusion process itself.

\section{Applications}
\label{applications}

\begin{figure*}[t]
    \centering
    \includegraphics[width=\textwidth]{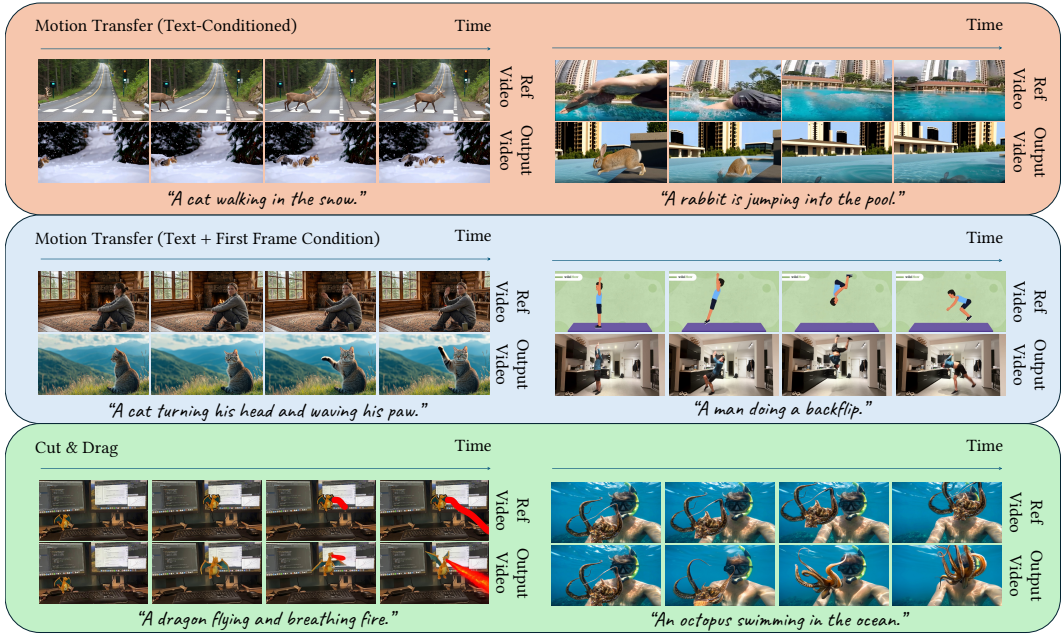}
    \caption{\textbf{Applications.} We showcase temporal conditioning under three settings: text-only conditioning (top), text combined with first-frame conditioning (middle), and Cut \& Drag inputs (bottom). In the middle and bottom rows, the first-frame condition is indicated by the leftmost frame in each sequence. (We recommend zooming in for a better view).}
    \label{fig: results}
\end{figure*}

\vspace{-0.25cm}
To demonstrate the capabilities of $\phi$-Noise, we present three applications under a single framework: text-conditioned motion transfer, text + first-frame motion transfer, and Cut \& Drag generation. Results are shown in \cref{fig: results}, Appendix B and the supplemental video. Further comparisons are provided in \cref{fig:comparisons}. We employ WAN~\cite{wan2025wanopenadvancedlargescale} for all of the following experiments and demonstrate additional results on LTX2\cite{ltx} in Appendix B.

\vspace{-0.25cm}
\paragraph{Text-Conditioned Motion Transfer}
Given a reference video and a text prompt, our goal is to generate a video that matches the prompt while preserving the input motion. As discussed in \cref{sec:analysis}, applying $\phi$-Noise along the spatial dimensions yields the best performance. The results demonstrate strong alignment with both the textual content and the motion patterns. Some spatial information is also transferred, which, while not always desirable, helps maintain temporal consistency and is also observed in competing methods.

\vspace{-0.25cm}
\paragraph{Text + First Frame Motion Transfer}
The goal is to align with both a text prompt and a first-frame condition while following the motion of the input video. We combine the WAN first-frame baseline with $\phi$-Noise. Our method successfully transfers motion across varying subjects (e.g., replacing a human with a cat) and handles complex dynamics, such as backflips, while preserving coherence.

\vspace{-0.25cm}
\paragraph{Cut-and-Drag Manipulations}
In this setting, users either cut object patches from an image or add an outside sprite on top of a given image, and animate them by dragging them rigidly across the frame. The goal is to generate a coherent video that follows the prescribed motion. We employ WAN with first-frame conditioning together with $\phi$-Noise.

As shown in \cref{fig: results}, our method produces natural object motion despite the rigid inputs. In the left example, it also introduces plausible visual effects (e.g., a red line becoming fire). The right example highlights the method’s flexibility: the octopus is not constrained to its rigid patch and instead moves freely while still adhering to the specified motion.

\vspace{-0.1cm}

\vspace{-0.25cm}
\section{Experiments}
\label{sec:experiments}

\vspace{-0.25cm}
To evaluate our spectral manipulation framework, we conduct extensive experiments across our three primary applications. We benchmark our approach against state-of-the-art diffusion guidance and motion transfer techniques, providing quantitative evaluations for Text-to-Video (T2V) and Cut \& Drag (CND) tasks, alongside qualitative comparisons across all settings.

\vspace{-0.25cm}
\subsection{Experimental Setup}

\vspace{-0.25cm}
\paragraph{Implementation Details.}
For all experiments, we apply our proposed $\phi$-Noise manipulation directly to the initial Gaussian noise prior to the diffusion denoising process of the Wan2.2-14B model~\cite{Wang2025WanOA}. For Image-to-Video (I2V) and Cut \& Drag tasks, we empirically select the frequency cutoff parameter $k \in [1,5]$ and fix the scaling coefficient to $\gamma = 30$. For the implicit temporal conditioning (T2V) task, we set $\gamma = 4$ and define $k$ as a continuous masking ratio, typically set to $\sim 5\%$.

\useunder{\uline}{\ul}{}

\begin{table}[ht]
\centering
\small
\renewcommand{\arraystretch}{1.3} %
\resizebox{\columnwidth}{!}{%
\begin{tabular}{llcccl|lccccc}
\hline
\textbf{}                                                       & \multirow{2}{*}{\textbf{Model}} & \multicolumn{3}{c}{\textbf{Image-Based Metrics}}         &  &  & \multicolumn{5}{c}{\textbf{Motion-Based Metrics}}                                                          \\ \cline{3-12} 
                                                                &                                 & CLIP-T$\ \uparrow$   & Aes$\ \uparrow$ & Img$\ \uparrow$ &  &  & LPIPS-T$\ \downarrow$ & Flow-E$\ \downarrow$ & Subj-C$\ \uparrow$ & Smooth$\ \uparrow$ & Dyn-D$\ \uparrow$ \\ \hline
\multirow{4}{*}{\rotatebox[origin=c]{90}{\textit{Cut \& Drag}}} & Wan-I2V                         & 0.308                & \textbf{0.652}  & {\ul 0.644}    &  &  & 0.116                 & 181.10               & 0.942              & {\ul 0.978}       & 0.647             \\
                                                                & GWTF                            & \textbf{0.314}       & 0.620           & 0.637           &  &  & \textbf{0.097}        & 152.81               & {\ul 0.942}       & \textbf{0.981}     & 0.647             \\
                                                                & TTM                             & 0.311                & {\ul 0.647}    & \textbf{0.653}  &  &  & {\ul 0.110}          & {\ul 102.39}        & \textbf{0.948}     & 0.978              & {\ul 0.705}      \\
                                                                & \textbf{Ours}                   & {\ul 0.313}         & 0.637           & 0.627           &  &  & 0.171                 & \textbf{101.49}      & 0.918              & 0.964              & \textbf{0.764}    \\ \hline
\multirow{5}{*}{\rotatebox[origin=c]{90}{\textit{T2V MT}}}      & Wan-T2V                         & 0.312                & \textbf{0.604}  & \textbf{0.705}  &  &  & \textbf{0.062}        & 103.26               & \textbf{0.955}     & {\ul 0.979}        & 0.645             \\
                                                                & DiT-Flow                           & {\ul \textbf{0.319}} & 0.526           & 0.611           &  &  & 0.112                 & 94.60                & 0.931              & 0.973              & \textbf{0.935}    \\
                                                                & DMT                             & {\ul 0.314}          & 0.530           & 0.581           &  &  & 0.114                 & {\ul 67.23}          & 0.914              & 0.963              & 0.871             \\
                                                                & MotionClone                         & 0.304                & 0.548           & 0.646           &  &  & 0.204                 & 67.92                & 0.864              & 0.919              & {\ul 0.903}      \\
                                                                & \textbf{Ours}                   & 0.302                & {\ul 0.546}    & {\ul 0.683}    &  &  & {\ul 0.075}          & \textbf{61.75}       & {\ul 0.952}       & \textbf{0.980}     & 0.709             \\ \hline
\end{tabular}

}
\vspace{0.1em}
\caption{\textbf{Quantitative Evaluation.} We report both Task-Specific Motion metrics and VBench~\cite{zheng2025vbench2} Generative Quality metrics across two tasks: Cut \& Drag and T2V Motion Transfer (T2V~MT). \textbf{Bold} and {\ul underline} indicate the best and second best performance among conditional manipulation methods, respectively. Wan-T2V/I2V serve as the unconditioned base models.
}
\label{tab:comprehensive_metrics}
\end{table}

\vspace{-0.35cm}
\paragraph{Datasets and Evaluation Benchmark.} 
We compile a diverse evaluation suite of 60 high-quality videos to rigorously assess motion transfer and structural preservation. This benchmark comprises 20 published examples from the Time-to-Move (TTM) dataset~\cite{time2move}, 30 videos sourced from the LOVEU-TGVE-2023 dataset~\cite{wu2023cvpr} (utilized specifically for evaluating object replacement captions), and 10 in-the-wild videos collected to test generalization on complex real-world dynamics. 

\vspace{-0.3cm}
\paragraph{Baselines.}
We employ the foundational Wan models (Wan-T2V and Wan-I2V) as our primary unconditioned baselines to establish standard text and image capabilities. For conditional generation, we compare against recent state-of-the-art approaches. For text-based generation, we compare against DiT-Flow~\cite{pondaven2025ditflow}, the T2V version of MotionClone~\cite{ling2024motionclone} and DMT~\cite{space_time}. For Cut \& Drag, we evaluate against Go-With-The-Flow (GWTF)~\cite{go_with_the_flow} and Time-to-Move (TTM)~\cite{time2move} and for TI2V Motion Transfer, we compare to IT2V MotionClone~\cite{ling2024motionclone} and I2V Wan 2.2 baseline~\cite{wan2025wanopenadvancedlargescale}. 

\vspace{-0.25cm}
\paragraph{Evaluation Metrics.}
We evaluate the generated videos across two categories. 
(1) \textbf{Task-Specific Metrics:} We measure dense motion alignment using \textbf{Flow-Err} (optical flow error between the reference and generated video), temporal consistency via \textbf{LPIPS-Temp}~\cite{zhang2018perceptual}, and semantic text alignment via \textbf{CLIP-T}~\cite{radford2021learning}. 
(2) \textbf{Generative Quality:} We utilize VBench~\cite{zheng2025vbench2}, a comprehensive evaluation suite for video diffusion, reporting Subject Consistency, Background Consistency, Motion Smoothness, Dynamic Degree, Aesthetic Quality, and Imaging Quality.

\begin{figure*}[t]
    \centering
    \includegraphics[width=.9\textwidth]{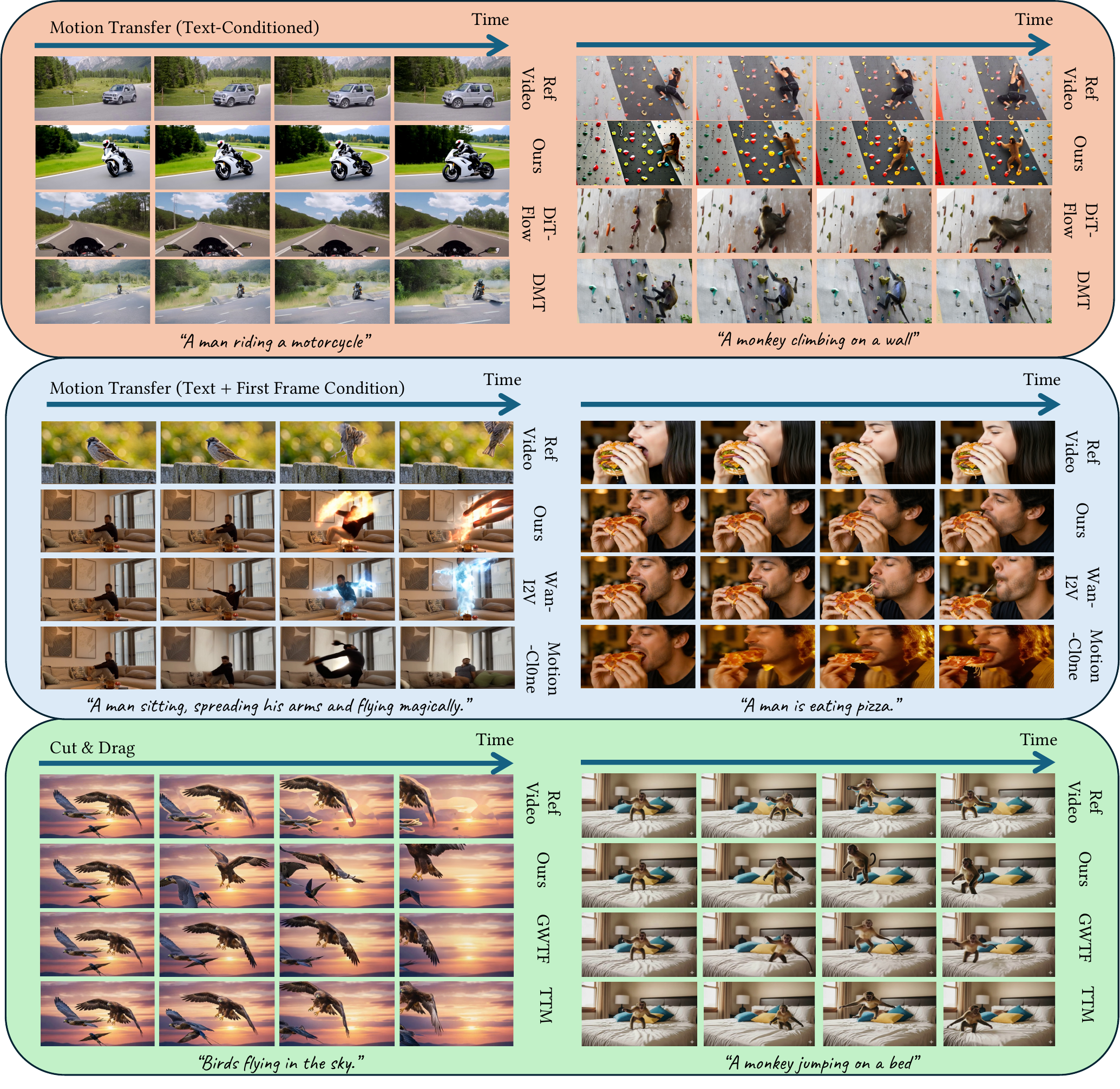}
    \caption{\textbf{Qualitative Comparisons.} We compare $\phi$-Noise with recent state-of-the-art methods for each application. In the middle and bottom rows, the first-frame condition is indicated by the leftmost frame in each sequence. (We recommend zooming in for a better view).}
    \vspace{-0.25cm}
    \label{fig:comparisons}
\end{figure*}

\vspace{-0.25cm}
\subsection{Qualitative Comparisons.}
\vspace{-0.25cm}

As shown in \cref{fig:comparisons}, in the text-only setting (top), $\phi$-Noise achieves strong motion transfer, with slight spatial leakage as the manipulation is applied along spatial dimensions. (see \cref{sec:analysis}). DiT-Flow exhibits weaker and sometimes missing motion transfer, while DMT improves alignment but introduces spatial leakage and remains inferior overall.

For text + first-frame conditioning, the Wan baseline fails to capture motion without explicit conditioning, whereas our method successfully produces complex motions aligned with the reference. MotionClone shows limited capability in this regard.

In the Cut \& Drag setting, GWTF yields stiff motion and visible artifacts. Both TTM and our method perform well, but differ in behavior: constrained by its mask condition, TTM adheres closely to the input patches, while our method produces more natural and expressive motion. This is evident in the bird example, where our results include realistic wing flapping, whereas TTM remains more constrained to the patch motion.

\vspace{-0.35cm}
\subsection{Quantitative Comparisons.}
\vspace{-0.25cm}
\paragraph{Motion Transfer and Consistency.}
\Cref{tab:comprehensive_metrics} summarizes the performance of our method against the baselines. Our approach demonstrates superior performance in dense motion alignment, achieving the lowest \textbf{Flow-Err} in both evaluated settings (T2V: 61.75, CND: 101.49). This validates our hypothesis from \cref{sec:analysis} that low-frequency phase directly dictates the global motion trajectories of the video.

\vspace{-0.25cm}
\paragraph{Generative Quality (VBench).}
A common limitation of guidance-based manipulation is the degradation of the model's native generative prior (Wan-T2V/I2V). However, by strictly conserving spectral energy, our formulation ensures the modified noise remains within the expected Gaussian distribution. In the T2V setting, our framework achieves the highest Subject Consistency (0.952) and Motion Smoothness (0.980) among all conditional baselines. In the CND setting, our method achieves the highest Dynamic Degree (0.764) while maintaining the lowest Flow-Err, proving that $\phi$-Noise synthesizes highly dynamic, accurate motion without compromising visual fidelity. We refer the reader to the Supplementary material to further illustrate our method's fidelity and visual quality, with additional qualitative comparisons.

\vspace{-0.25cm}
\paragraph{Computational Efficiency.} Unlike test-time optimization or attention-injection methods that introduce heavy per-step overhead during denoising, $\phi$-Noise is highly efficient. By modifying only the initial noise $\mathbf{z}$ via a single FFT operation, it introduces near-zero latency. Consequently, our method maintains the same inference latency as the Wan2.2 baseline.

\subsection{Additional Experiments}
\vspace{-0.25cm}
We conduct additional experiments to further evaluate the capabilities of $\phi$-Noise. These include seed variation analysis, ablations over the choice of $\gamma$ and $k$, prompt variation experiments, applying $\phi$-Noise to an additional video generation model, and extending it to image generation models. We present these experiments, along with further comparisons, in Appendix B. 
\vspace{-0.1cm}

\vspace{-0.25cm}
\section{Limitations and Conclusion}
\label{conclusion}

\vspace{-0.25cm}

In this paper, we introduced $\phi$-Noise, a simple and efficient cross-task framework for motion transfer based on manipulating the low-frequency phase components of the input noise in the Fourier domain. Through extensive analysis, we showed that directly modifying the noise is non-trivial, as it disrupts the spectral balance of the latent signal. To address this issue, we proposed an energy-balancing mask that `re-whitens' the manipulated Gaussian latent prior to denoising, keeping it aligned with the expected distribution of the generative model.

Our method’s primary limitation lies in its sensitivity to the parameter space, particularly the masking ratio $k$. While improper tuning can lead to structural artifacts or motion mistransfer, it also provides meaningful control over the degree of the transferred motion.

We evaluated $\phi$-Noise across multiple motion transfer tasks, demonstrating strong performance compared to prior training- and optimization-based approaches, which are often tailored to specific settings. More broadly, our results highlight input noise as a powerful and underexplored conditioning space, suggesting that frequency-based noise manipulation can serve as a general and flexible framework for controllable video generation.

\clearpage
\bibliographystyle{plainnat}
\bibliography{neurips_2026}

\clearpage

\appendix

\noindent{\Large \bf Appendix \par}

\section{Derivation of the Energy-Balanced Compensation Factor}

To maintain spectral consistency during phase mixing, we define a Balanced-Energy Mask. Given the latent noise $\mathbf{\mathbf{\tilde{z}}}$ and parameters $\gamma, k$, we scale the $k$ lowest frequency components by $1/\gamma$ while preserving the total signal energy $\text{E}(\mathbf{\mathbf{\tilde{z}}})$.

\subsection*{Energy Decomposition}
We partition the total energy $\text{E}(\mathbf{\tilde{z}})$ into low-frequency ($\text{E}_{\text{l}}$) and high-frequency ($\text{E}_{\text{h}}$) components based on the threshold index $k$:
\begin{equation}
    \text{E}_{\text{l}} = \sum_{i=0}^{k} |\mathbf{\mathbf{\tilde{z}}}_i|^2, \quad \text{E}_{\text{h}} = \sum_{i=k+1}^{t-1} |\mathbf{\mathbf{\tilde{z}}}_i|^2
\end{equation}
where $\text{E}(\mathbf{\mathbf{\tilde{z}}}) = \text{E}_{\text{l}} + \text{E}_{\text{h}}$ by the additivity of the squared Frobenius norm.

\subsection*{Conservation Constraint}
Let $\mathbf{\tilde{z}}_k$ denote the energy-balanced noise. We scale the low-frequency components by $\frac{1}{\gamma}$ and the high-frequency components by a compensation factor $\beta$. We require:
\begin{equation}
    \text{E}(\mathbf{\mathbf{\tilde{z}}}) = \text{E}(\mathbf{\mathbf{\tilde{z}}}_k)
\end{equation}

Expanding the energy of the modified signal:
\begin{align}
    \text{E}(\mathbf{\mathbf{\tilde{z}}}_k) &=  \sum_{i=0}^{k} |\frac{1}{\gamma}\cdot\mathbf{\mathbf{\tilde{z}}}_i|^2 + \sum_{i=k+1}^{t-1} |\beta\cdot\mathbf{\mathbf{\tilde{z}}}_i|^2 \\
    &= \left(\frac{1}{\gamma}\right)^2\cdot\sum_{i=0}^{k} |\mathbf{\mathbf{\tilde{z}}}_i|^2 + \beta^2\cdot\sum_{i=k+1}^{t-1} |\mathbf{\mathbf{\tilde{z}}}_i|^2 \\&= \frac{1}{\gamma^2} \text{E}_{\text{l}} + \beta^2 \text{E}_{\text{h}}
\end{align}

\subsection*{Closed-form Expression for $\beta$}
Equating the terms to satisfy the energy conservation constraint:
\begin{equation}
    \text{E}(\mathbf{\mathbf{\tilde{z}}}) = \frac{1}{\gamma^2} \text{E}_{\text{l}} + \beta^2 \text{E}_{\text{h}}
\end{equation}

Rearranging to isolate $\beta$:
\begin{equation}
    \beta^2\cdot\text{E}_{\text{h}} = \text{E}(\mathbf{\mathbf{\tilde{z}}}) - \frac{\text{E}_{\text{l}}}{\gamma^2}
\end{equation}

Which yields the final expression:
\begin{equation}
    \beta = \sqrt{\frac{\text{E}(\mathbf{\mathbf{\tilde{z}}}) - \frac{\text{E}_{\text{l}}}{\gamma^{2}}}{\text{E}_{\text{h}}}}.
\end{equation}

\section{Additional Experiments}
\subsection{$\phi$-Noise for Image Generation}
\begin{figure*}[t]
    \centering
    \includegraphics[width=\textwidth]{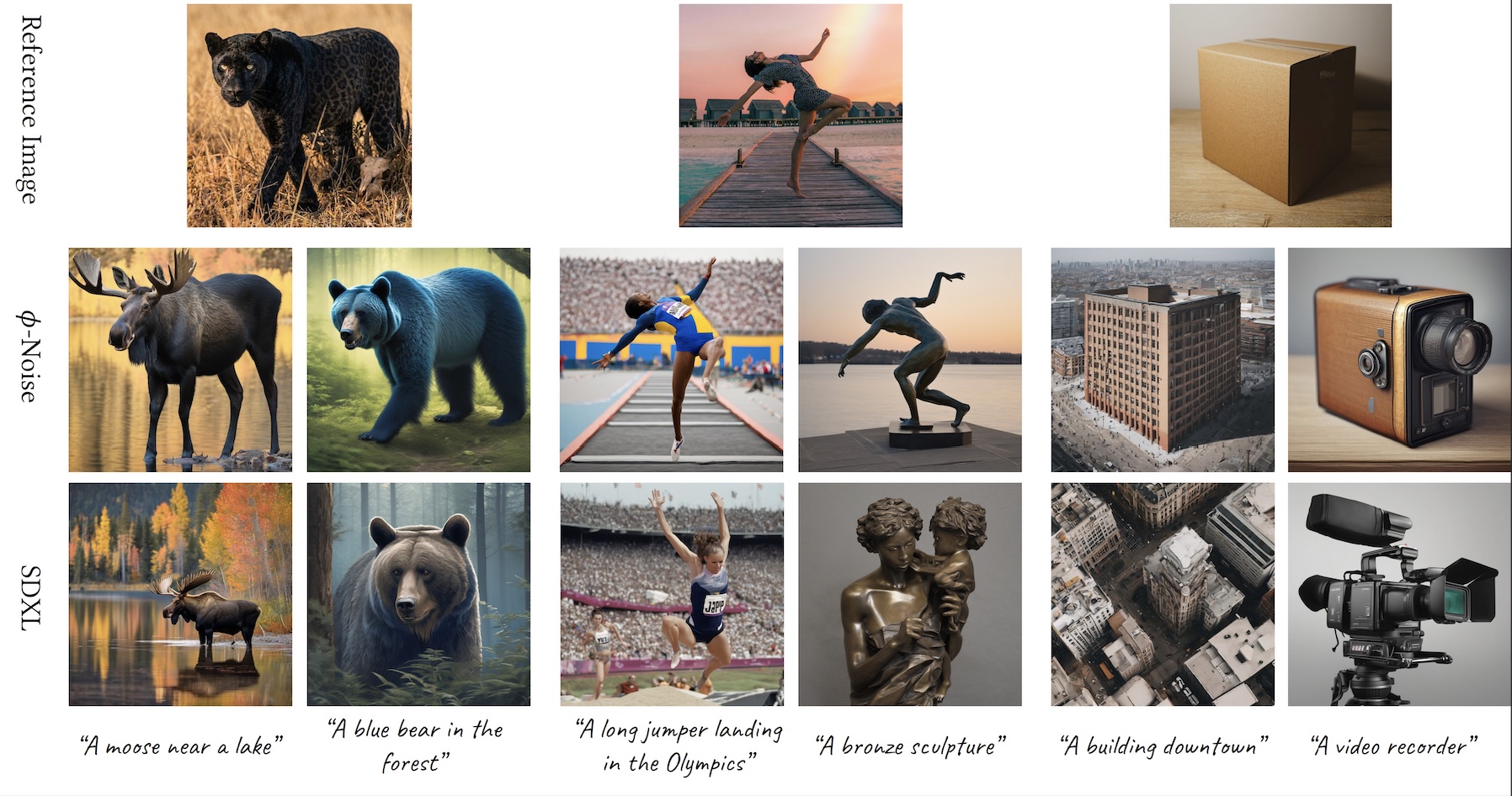}
    \caption{\textbf{$\phi$-Noise for Image Generation.} We apply $\phi$-Noise to SDXL by injecting spatial phase information into the input noise. As shown, the biased noise enables the generated images (middle row) to spatially align with the reference image (top row), whereas generation with unbiased noise (bottom row) exhibits different spatial arrangements and alignment patterns.}
    \label{fig:phi_image}
\end{figure*}

As shown in Section 3 of the main manuscript, $\phi$-Noise can be applied in the spatial domain to preserve motion cues from an input video. In this section, we demonstrate that the same principle can also be extended to image generation models. Specifically, we employ SDXL~\cite{podell2023sdxlimprovinglatentdiffusion} and spatially bias its input noise using $\mathcal{F}_{\text{S}}$, comparing the resulting outputs to those generated by vanilla SDXL across a variety of prompts and reference images. Results are presented in \cref{fig:phi_image}.

As observed, the biased outputs (middle row) exhibit spatial structures similar to those of the reference images (top row), effectively enabling loose pose conditioning without requiring additional training or significant inference overhead, unlike prior approaches such as~\cite{control_net}. Notably, our method generalizes across a wide range of object categories, including animals, humans, and even inanimate objects such as boxes, whereas many existing pose-conditioning methods are primarily limited to humans due to data and supervision constraints. In contrast, generation with unbiased noise fails to preserve the spatial alignment of the reference image, as expected.

\subsection{Model Generalization}
\begin{figure}[h]
\centering

\vspace{1mm}

\newlength{\rowheightT}
\setlength{\rowheightT}{0.095\textheight}

\newlength{\imheightT}
\setlength{\imheightT}{0.75\rowheightT}

\newlength{\labelwidthT}
\setlength{\labelwidthT}{0.03\linewidth}

\newcommand{\resultrowT}[3]{%
  \noindent
  \begin{minipage}[c][\imheightT][c]{\labelwidthT}
    \centering
    \rotatebox{90}{\small\bfseries #2}
  \end{minipage}%
  \hfill
  \begin{minipage}[c][\imheightT][c]{0.955\linewidth}
    \includegraphics[width=\linewidth,height=\imheightT]{#1}
  \end{minipage}
  \par\vspace{0.6mm}
}

{\small\fontfamily{Caveat-TLF}\selectfont
``Several people breakdance in a school gym.''\par}

\vspace{0.6mm}

\resultrowT{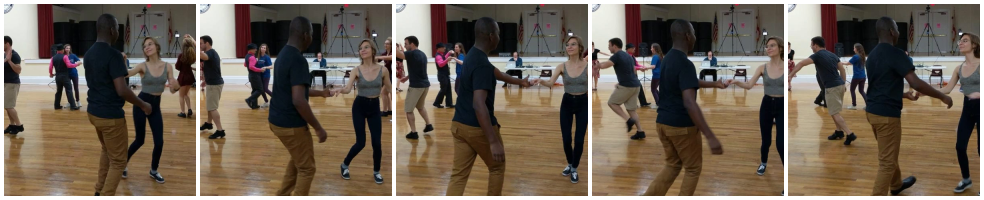}{Input}{}

\resultrowT{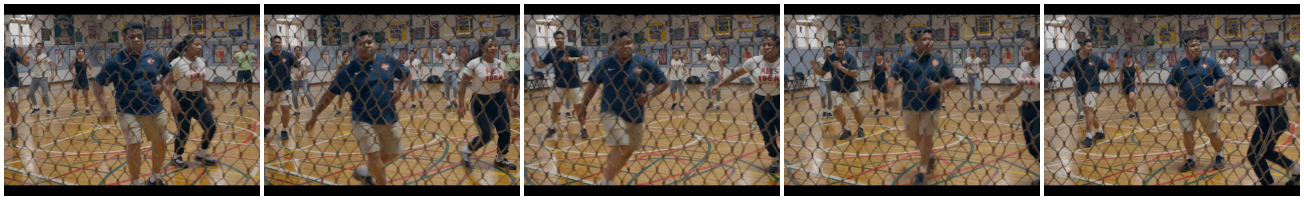}{Output}{}

\vspace{1.2mm}

{\small\fontfamily{Caveat-TLF}\selectfont
``A toy fairy floating and emits a bright red magical beam.''\par}

\vspace{0.6mm}

\resultrowT{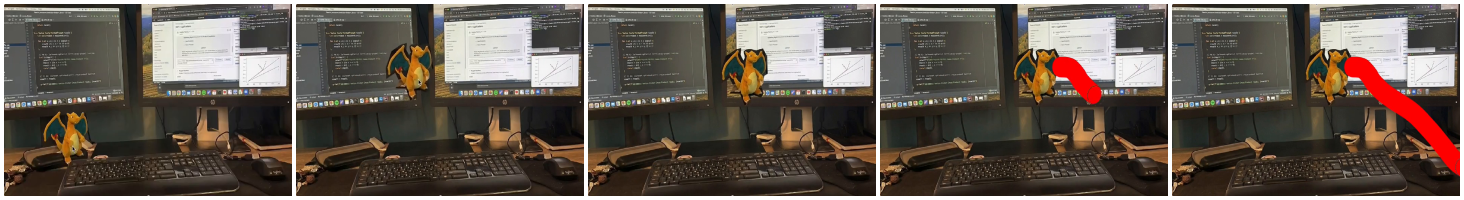}{Input}{}

\resultrowT{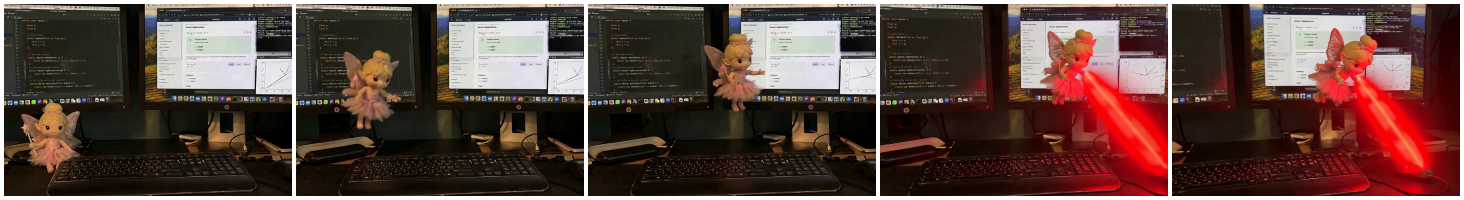}{Output}{}

\vspace{1.2mm}

{\small\fontfamily{Caveat-TLF}\selectfont
``A toy is skateborading and jupming in the air.''\par}

\vspace{0.6mm}

\resultrowT{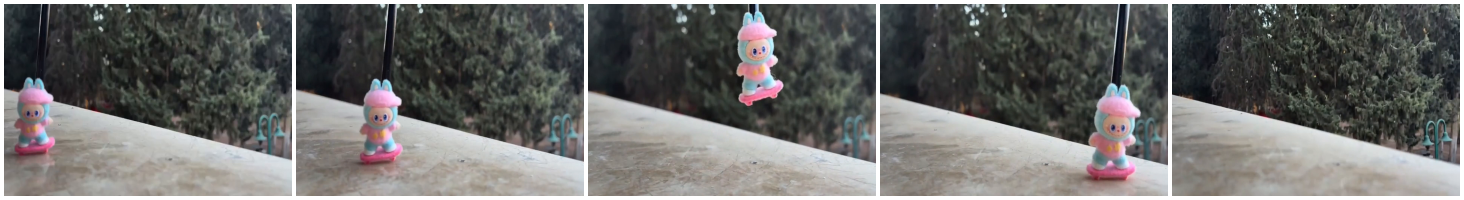}{Input}{}

\resultrowT{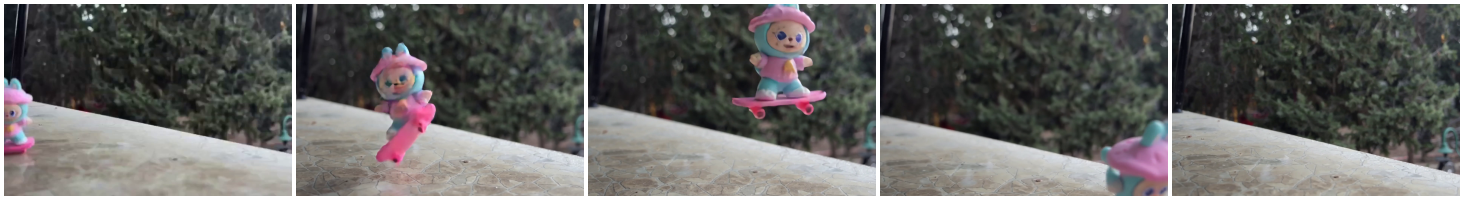}{Output}{}

\caption{
\textbf{Applications with LTX-based video generation.}
We demonstrate multiple applications using LTX text-to-video and image-to-video models. For each example, the first row shows the input and the second row shows the generated output conditioned on the corresponding prompt.
}
\label{fig:ltx_applications}
\end{figure}

To evaluate the generality of $\phi$-Noise across different architectures, we apply our method to LTX2~\cite{ltx}. We present results in \cref{fig:ltx_applications} for text-conditioned motion transfer (top), text + first-frame conditioning (middle), and Cut \& Drag inputs (bottom). As demonstrated, $\phi$-Noise can be effectively integrated with LTX2 and produces pleasing results across all evaluated applications.

\subsection{Seed Variations}
\begin{figure*}[t]
\centering

\vspace{1mm}

\setlength{\rowheightT}{0.105\textheight}

\setlength{\imheightT}{0.75\rowheightT}

\newcommand{\resultrowT}[1]{%
  \noindent
  \includegraphics[width=\linewidth,height=\imheightT]{#1}
  \par\vspace{1mm}%
}

\resultrowT{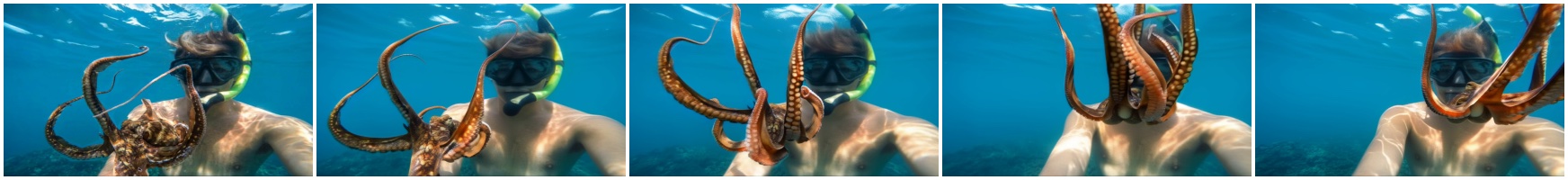}

\resultrowT{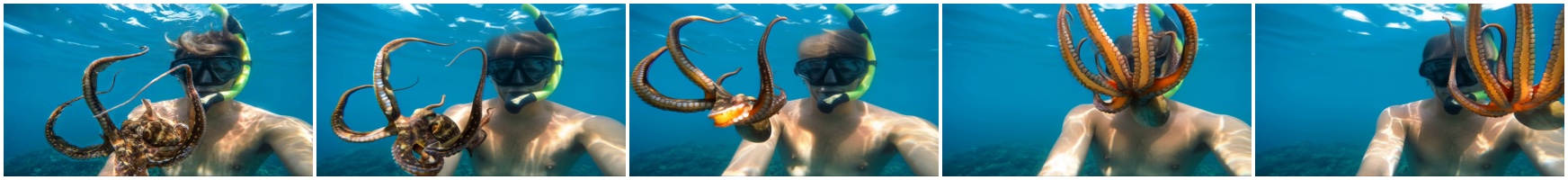}

\resultrowT{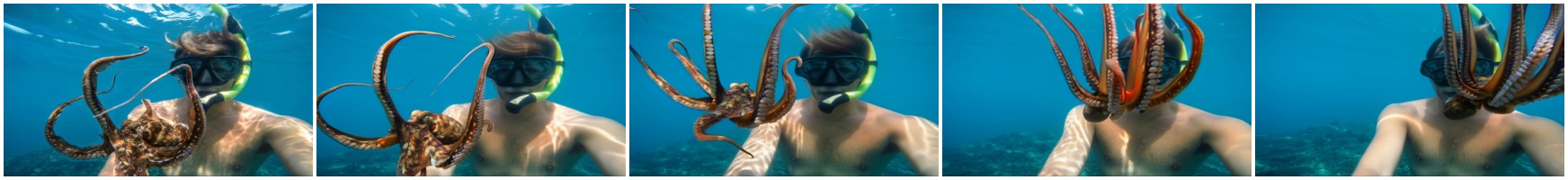}

\resultrowT{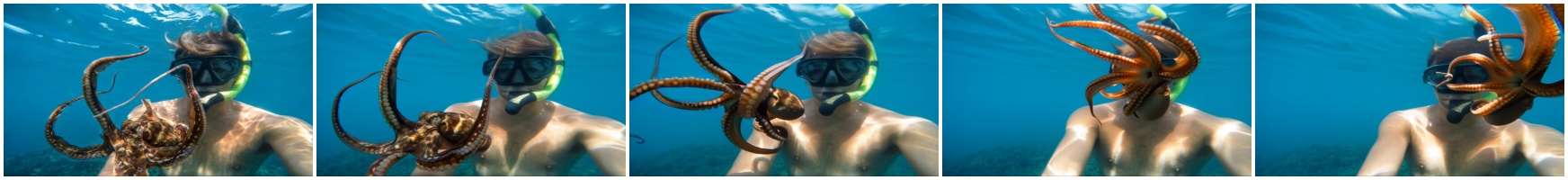}

\resultrowT{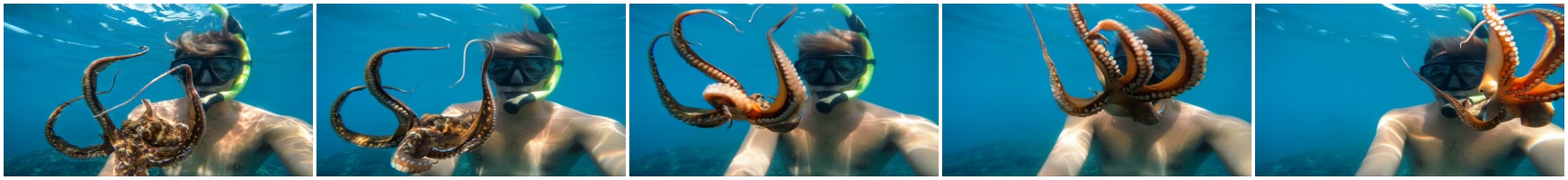}

\caption{
\textbf{Seed Variation.} Showcasing seed variation on a single video input for Cut \& Drag generation. Each row denotes a different random seed. Prompt: \textit{``An octopus swimming in the ocean.''}
}
\label{fig:sv0}
\end{figure*}

\begin{figure*}[t]
\centering

\vspace{1mm}

\setlength{\rowheightT}{0.105\textheight}

\setlength{\imheightT}{0.75\rowheightT}

\newcommand{\resultrowT}[1]{%
  \noindent
  \includegraphics[width=\linewidth,height=\imheightT]{#1}
  \par\vspace{1mm}%
}

\resultrowT{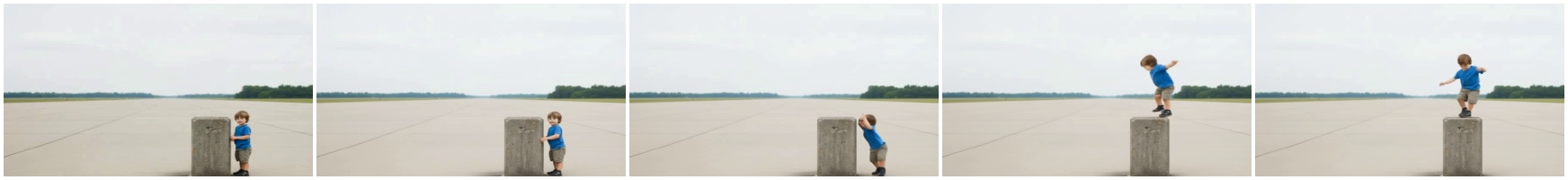}

\resultrowT{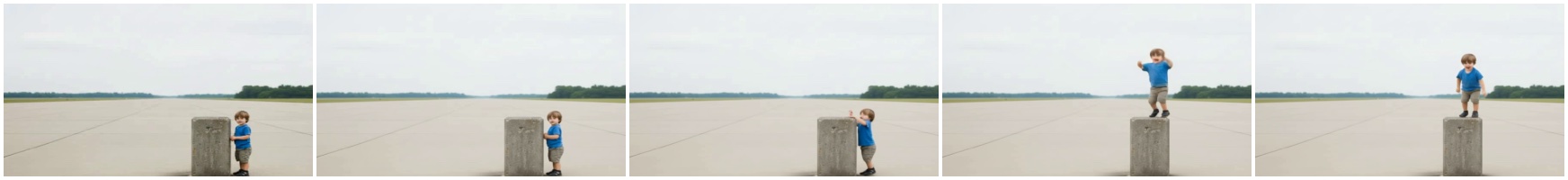}

\resultrowT{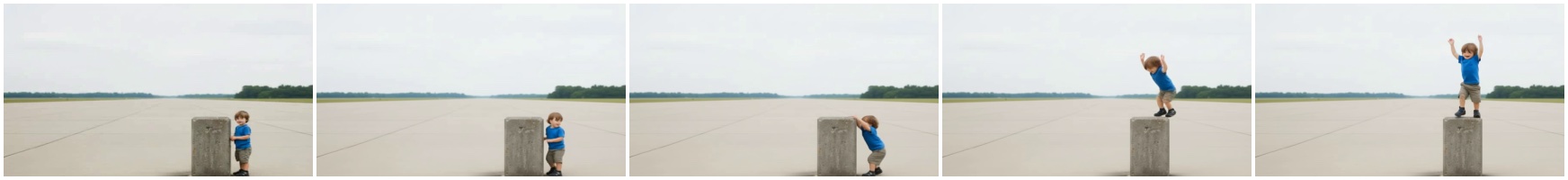}

\resultrowT{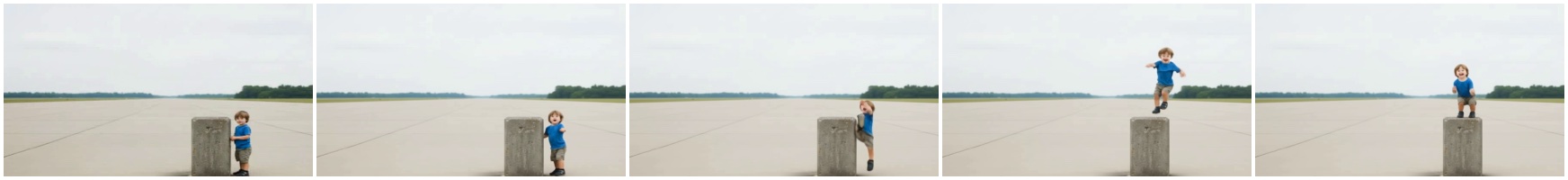}

\resultrowT{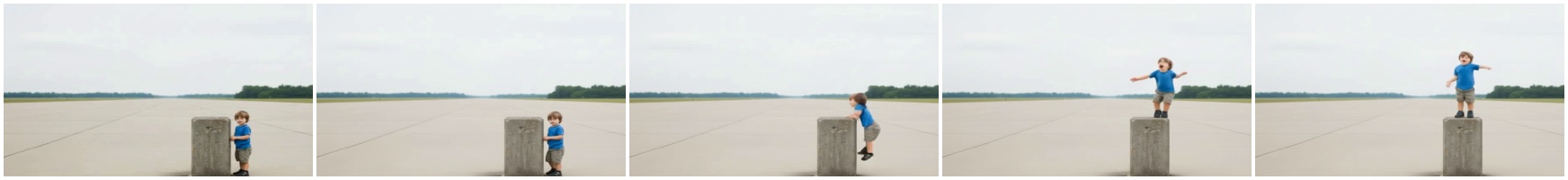}

\caption{
\textbf{Seed Variation.} Showcasing seed variation on a single video input for Cut \& Drag generation. Each row denotes a different random seed. Prompt: \textit{``A little boy jumping on a pillar''.}
}
\label{fig:sv1}
\end{figure*}

Since $\phi$-Noise conditions the noisy input prior to the diffusion process, the model retains the ability to modify and refine the output throughout generation. As a result, the generated samples exhibit slight variations across different random seeds, enabling exploration of diverse outputs while preserving the conditioning signal. We showcase seed variations in \cref{fig:sv0,fig:sv1}.

\subsection{Prompt Variations}
\begin{figure*}[t]
\centering

\vspace{1mm}

\setlength{\rowheightT}{0.105\textheight}

\setlength{\imheightT}{0.75\rowheightT}

\newcommand{\resultrowT}[2]{%
  \noindent
  {\Large\fontfamily{Caveat-TLF}\selectfont #2\par}
  \vspace{0.5mm}
  \includegraphics[width=\linewidth,height=\imheightT]{#1}
  \par\vspace{1mm}%
}

\resultrowT{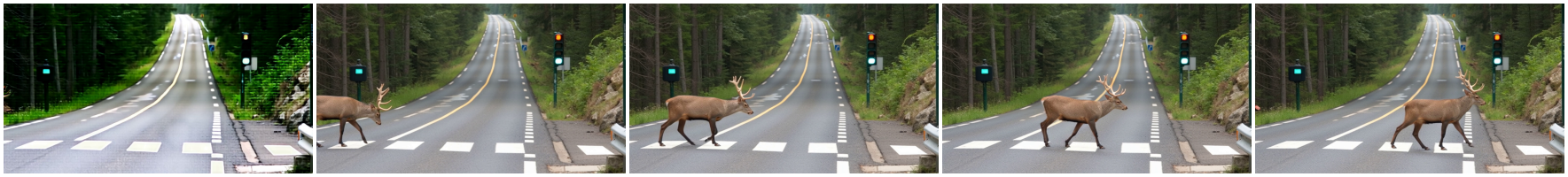}
{\small Reference Video.}

\resultrowT{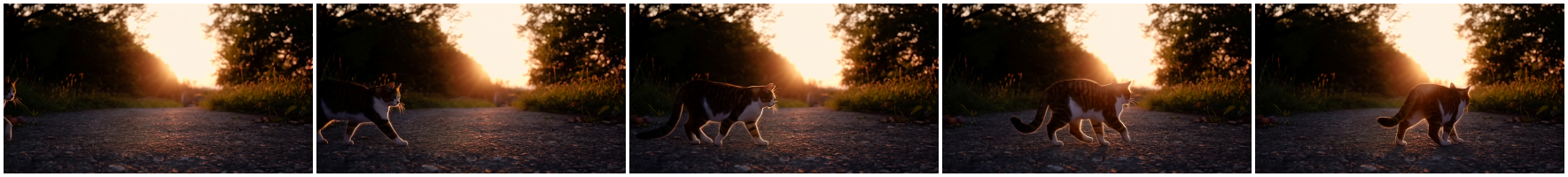}
{\small ``A cat is walking during sunset.''}

\resultrowT{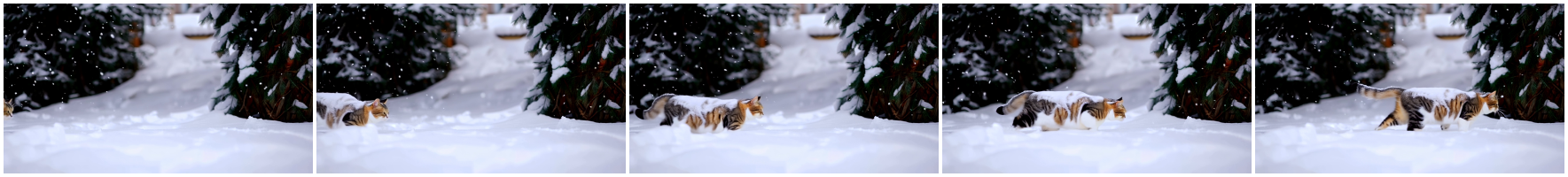}
{\small ``A cat is walking in the snow.''}

\resultrowT{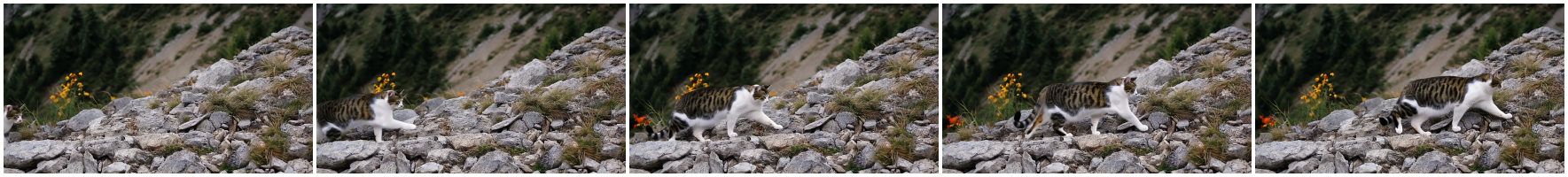}
{\small ``A cat is walking on a mountain.''}

\resultrowT{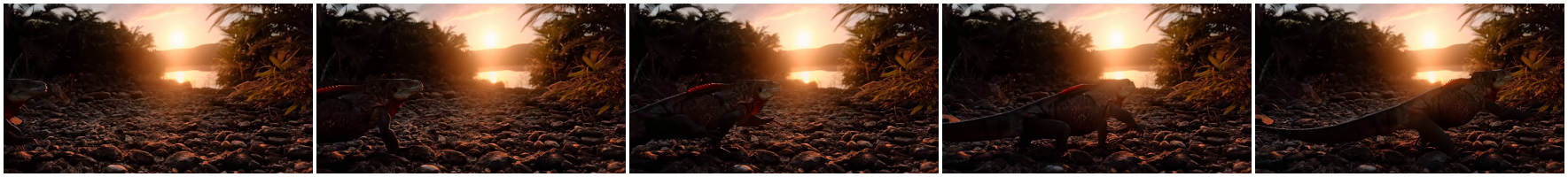}
{\small ``An iguana is walking during sunset.''}

\resultrowT{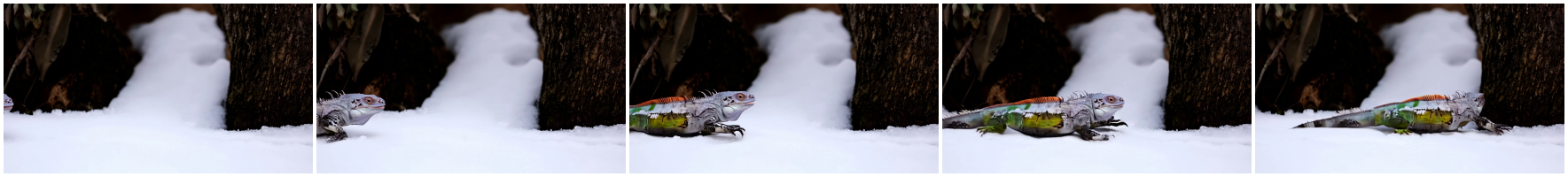}
{\small ``An iguana is walking in the snow.''}

\resultrowT{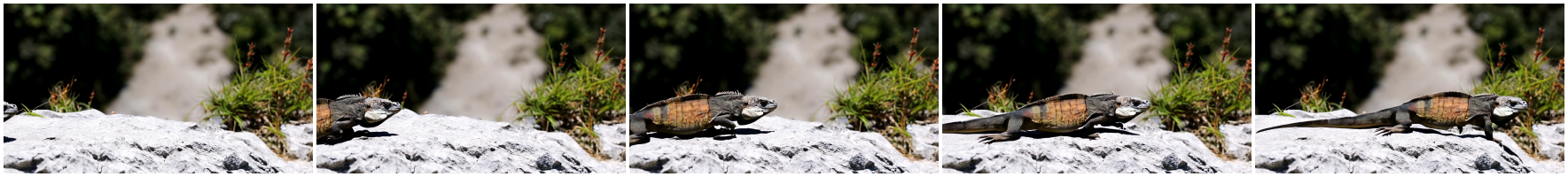}
{\small ``An iguana is walking on a mountain.''}

\caption{
\textbf{Prompt Variation.}
Given a reference video (top) we generate various videos depicting different animals and environments. As can be observed, all samples depict the reference video's motion.
}
\label{fig:prompt_var}
\end{figure*}

To demonstrate our method’s ability to adapt a single reference video to diverse prompts, we generate multiple videos with varying subjects and environments using the same reference. Results are shown in \cref{fig:prompt_var}.

\subsection{Comparisons}
\begin{figure*}[t]
\centering

{\Large\bfseries T2V Motion Transfer Comparison\par}
\vspace{1mm}

\setlength{\rowheightT}{0.105\textheight}

\setlength{\imheightT}{0.75\rowheightT}

\setlength{\labelwidthT}{0.035\linewidth}

\newcommand{\resultrowT}[2]{%
  \noindent
  \begin{minipage}[c][\imheightT][c]{\labelwidthT}
    \centering
    \rotatebox{90}{\small\bfseries #2}
  \end{minipage}%
  \hfill
  \begin{minipage}[c][\imheightT][c]{0.95\linewidth}
    \includegraphics[width=\linewidth,height=\imheightT]{#1}
  \end{minipage}%
  \par\vspace{0pt}%
}

\resultrowT{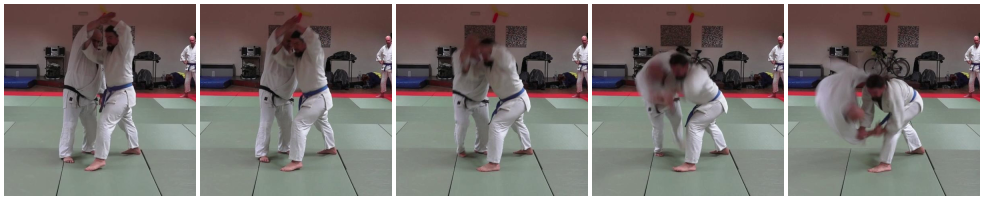}{Input}
\resultrowT{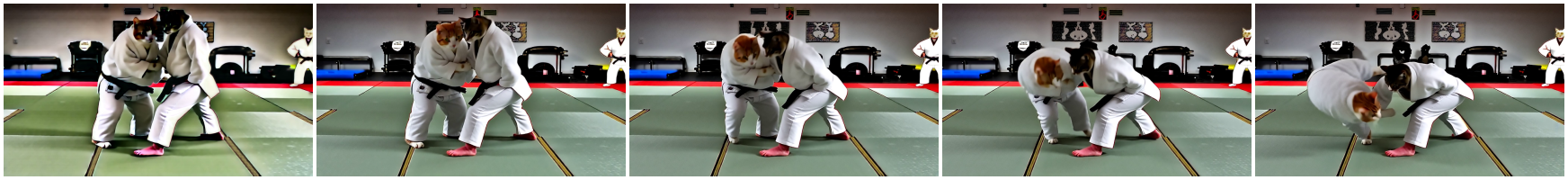}{Ours}
\resultrowT{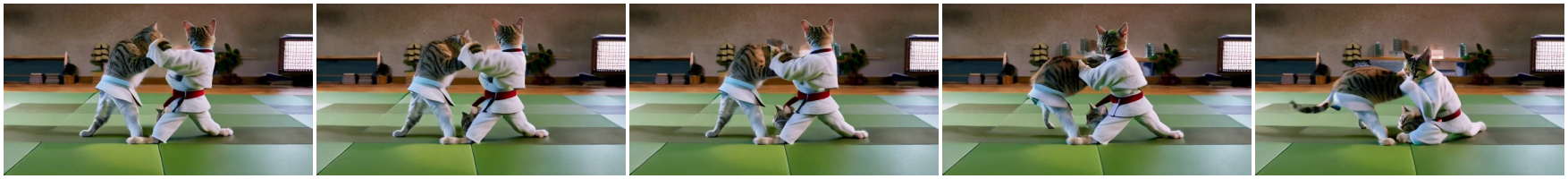}{DMT}
\resultrowT{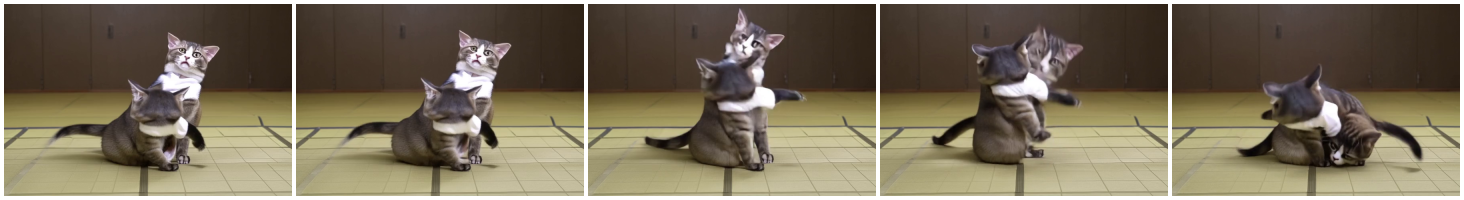}{DiTFlow}
\resultrowT{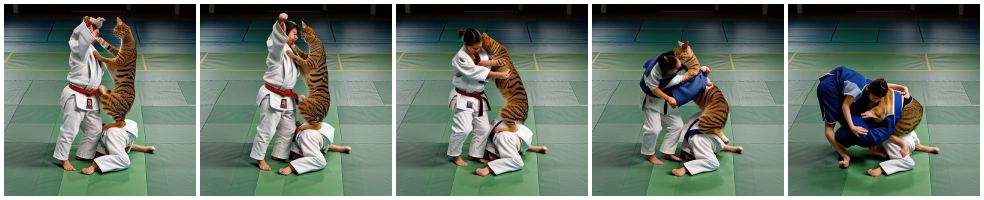}{MotionClone}
\resultrowT{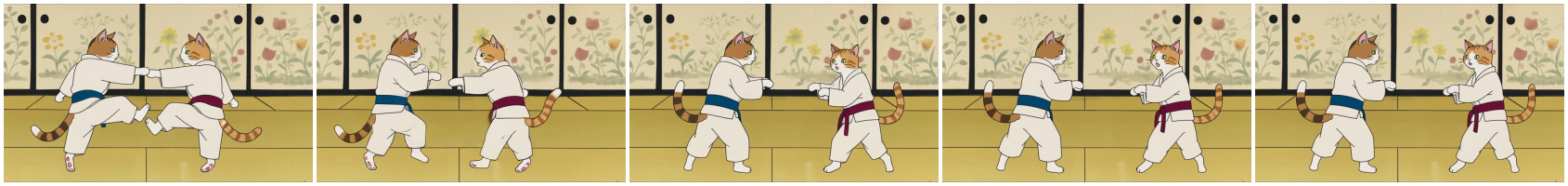}{Wan}

\caption{
\textbf{T2V Motion Transfer comparison.}
Qualitative comparison between different methods using the prompt
\text{``Two cats sparring in a dojo.''}
while preserving the motion dynamics from the input video.
}
\label{fig:c0}
\end{figure*}

\begin{figure*}[t]
\centering

{\Large\bfseries I2V Motion Transfer Comparison\par}
\vspace{1mm}

\newlength{\rowheightI}
\setlength{\rowheightI}{0.105\textheight}

\newlength{\imheightI}
\setlength{\imheightI}{0.75\rowheightI}

\newlength{\labelwidthI}
\setlength{\labelwidthI}{0.035\linewidth}

\newcommand{\resultrowI}[2]{%
  \noindent
  \begin{minipage}[c][\imheightI][c]{\labelwidthI}
    \centering
    \rotatebox{90}{\small\bfseries #2}
  \end{minipage}%
  \hfill
  \begin{minipage}[c][\imheightI][c]{0.95\linewidth}
    \includegraphics[width=\linewidth,height=\imheightI]{#1}
  \end{minipage}%
  \par\vspace{0pt}%
}

\resultrowI{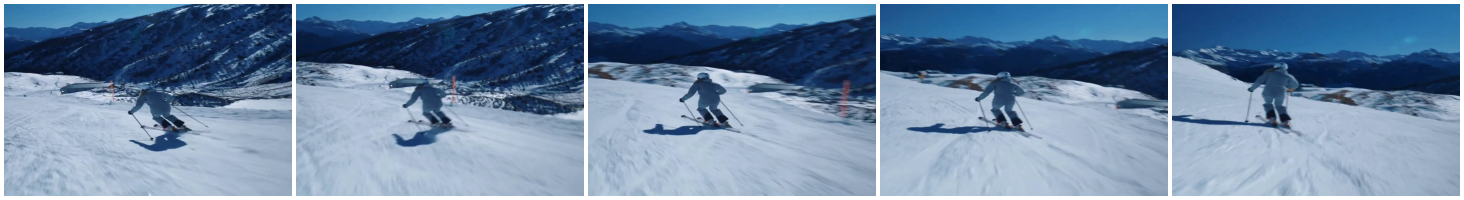}{Input}
\resultrowI{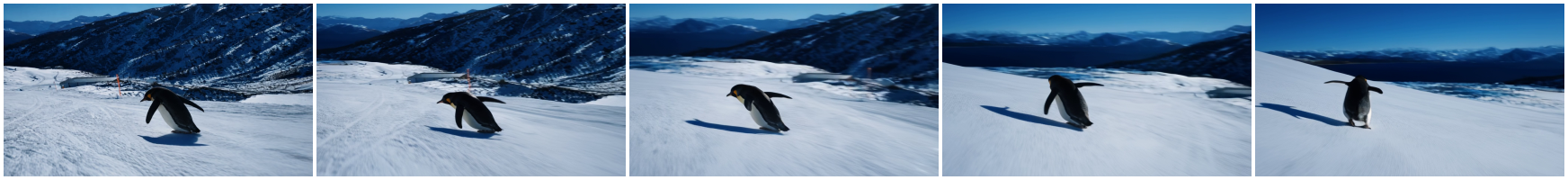}{Ours}
\resultrowI{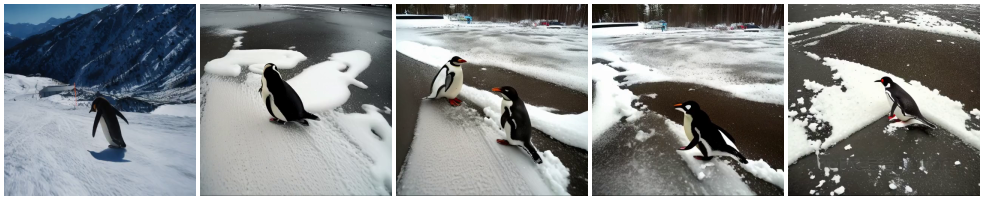}{MotionClone}
\resultrowI{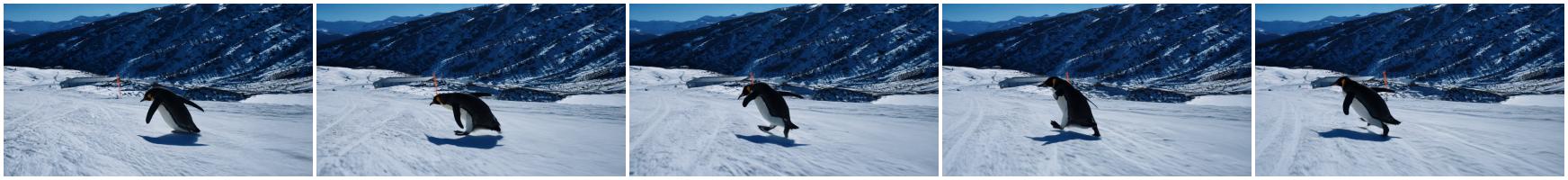}{Wan}

\caption{\textbf{I2V Motion Transfer comparison.}
Results generated from an input image and the text prompt
\textit{``A penguin sliding down a snowy slope.''}
while preserving the transferred motion dynamics. The first frame is shown in the left column.
}
\label{fig:c1}
\end{figure*}

\begin{figure*}[t]
\centering

{\Large\bfseries Cut \& Drag Comparison\par}
\vspace{1mm}

\setlength{\rowheightI}{0.105\textheight}

\setlength{\imheightI}{0.75\rowheightI}

\setlength{\labelwidthI}{0.035\linewidth}

\newcommand{\resultrowI}[2]{%
  \noindent
  \begin{minipage}[c][\imheightI][c]{\labelwidthI}
    \centering
    \rotatebox{90}{\small\bfseries #2}
  \end{minipage}%
  \hfill
  \begin{minipage}[c][\imheightI][c]{0.95\linewidth}
    \includegraphics[width=\linewidth,height=\imheightI]{#1}
  \end{minipage}%
  \par\vspace{0pt}%
}

\resultrowI{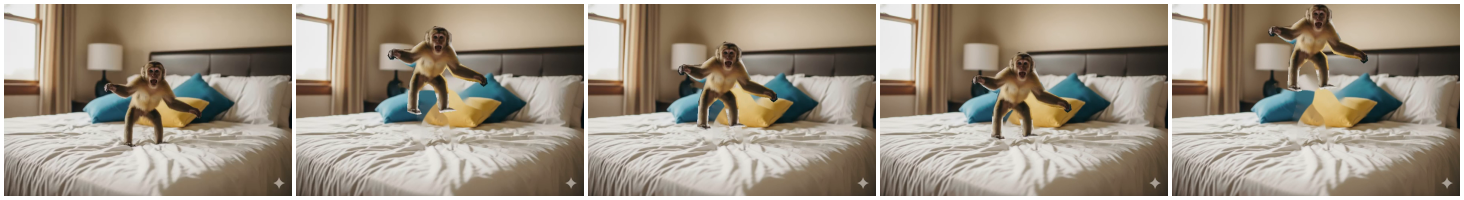}{Input}
\resultrowI{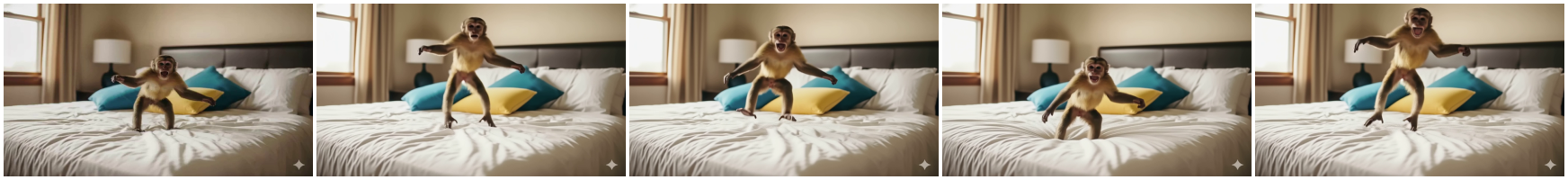}{Ours}
\resultrowI{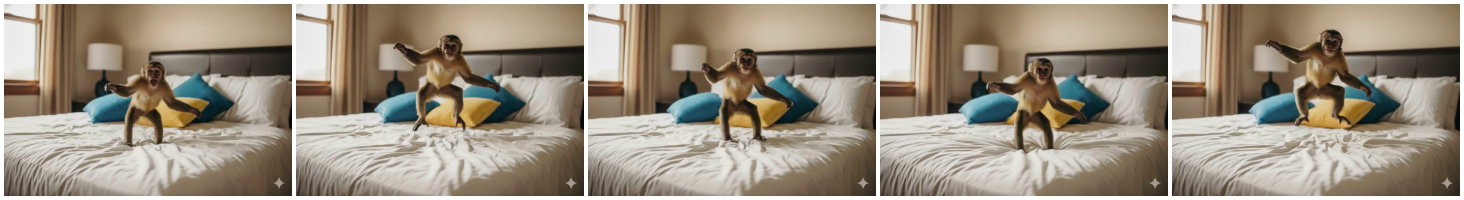}{ttm}
\resultrowI{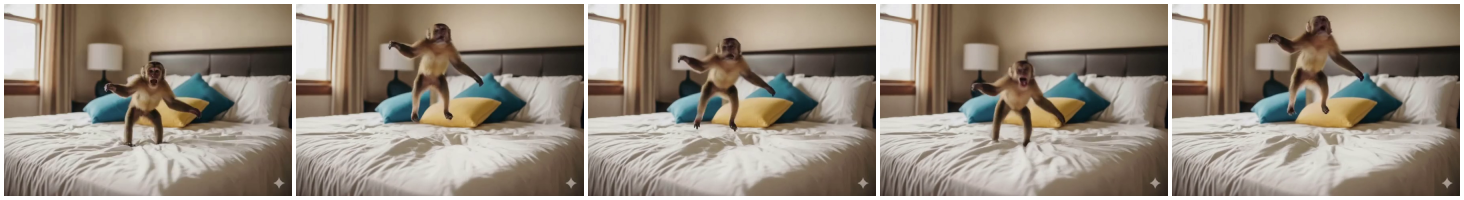}{GWTF}

\caption{\textbf{Cut \& Drag Comparison.}
Results generated from an input image and the text prompt
\textit{``A monkey jumping on the bed.''}
while preserving the transferred motion dynamics.
}
\label{fig:c2}
\end{figure*}

\begin{figure*}[t]
\centering

{\Large\bfseries T2V Motion Transfer Comparison\par}
\vspace{1mm}

\setlength{\rowheightT}{0.105\textheight}

\setlength{\imheightT}{0.75\rowheightT}

\setlength{\labelwidthT}{0.035\linewidth}

\newcommand{\resultrowT}[2]{%
  \noindent
  \begin{minipage}[c][\imheightT][c]{\labelwidthT}
    \centering
    \rotatebox{90}{\small\bfseries #2}
  \end{minipage}%
  \hfill
  \begin{minipage}[c][\imheightT][c]{0.95\linewidth}
    \includegraphics[width=\linewidth,height=\imheightT]{#1}
  \end{minipage}%
  \par\vspace{0pt}%
}

\resultrowT{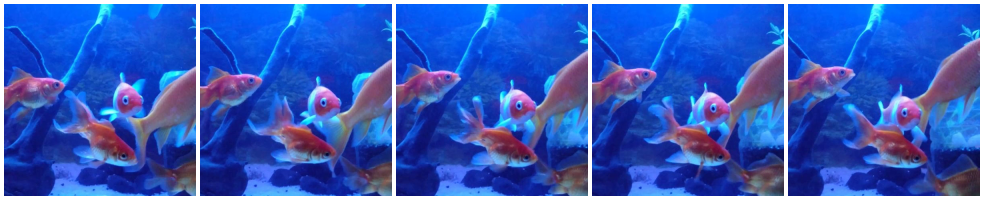}{Input}
\resultrowT{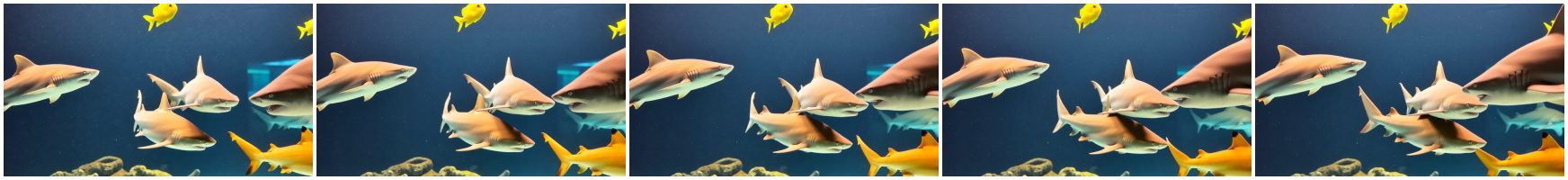}{Ours}
\resultrowT{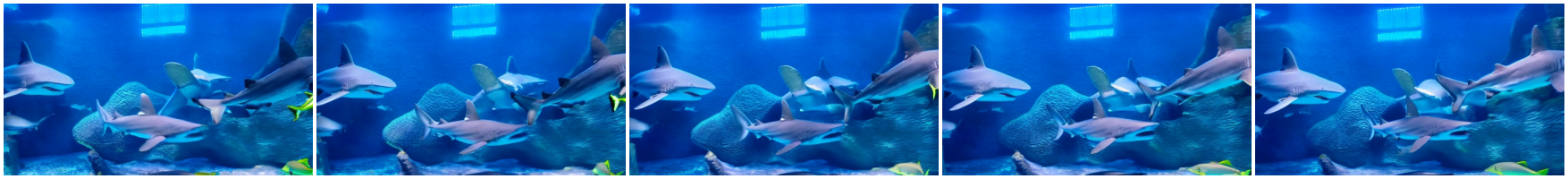}{DMT}
\resultrowT{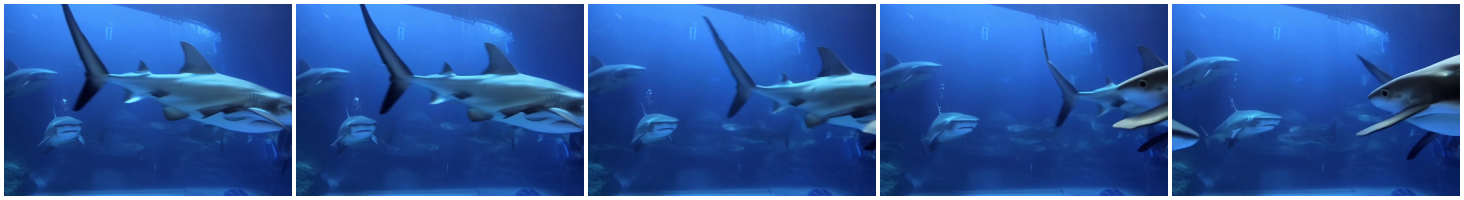}{DiTFlow}
\resultrowT{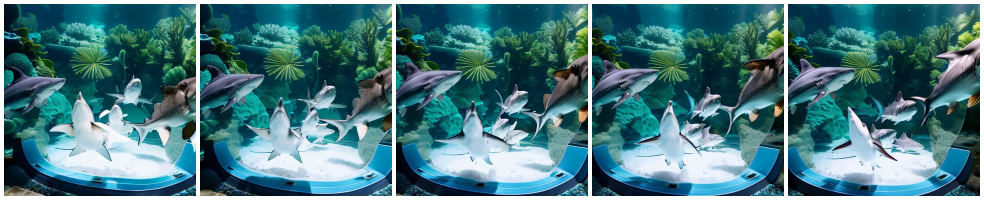}{MotionClone}
\resultrowT{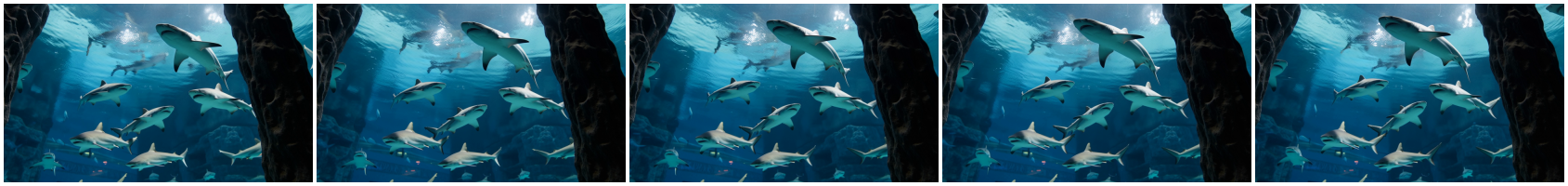}{Wan}

\caption{
\textbf{T2V Motion Transfer comparison.}
Qualitative comparison between different methods using the prompt
\textit{``Several sharks swim in a tank.''}
while preserving the motion dynamics from the input video.
}
\label{fig:c3}
\end{figure*}

\begin{figure*}[t]
\centering

{\Large\bfseries I2V Motion Transfer Comparison\par}
\vspace{1mm}

\setlength{\rowheightI}{0.105\textheight}

\setlength{\imheightI}{0.75\rowheightI}

\setlength{\labelwidthI}{0.035\linewidth}

\newcommand{\resultrowI}[2]{%
  \noindent
  \begin{minipage}[c][\imheightI][c]{\labelwidthI}
    \centering
    \rotatebox{90}{\small\bfseries #2}
  \end{minipage}%
  \hfill
  \begin{minipage}[c][\imheightI][c]{0.95\linewidth}
    \includegraphics[width=\linewidth,height=\imheightI]{#1}
  \end{minipage}%
  \par\vspace{0pt}%
}

\resultrowI{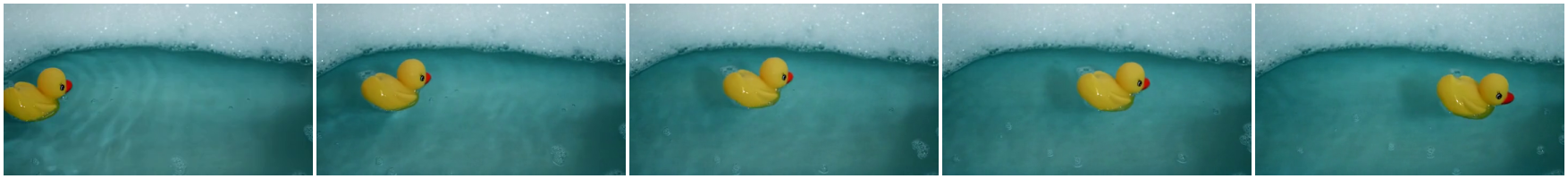}{Input}
\resultrowI{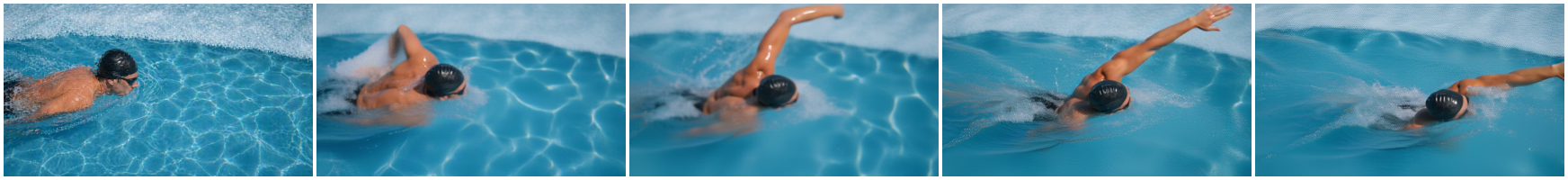}{Ours}
\resultrowI{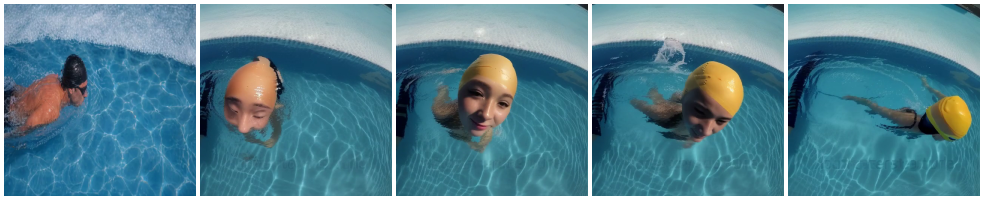}{MotionClone}
\resultrowI{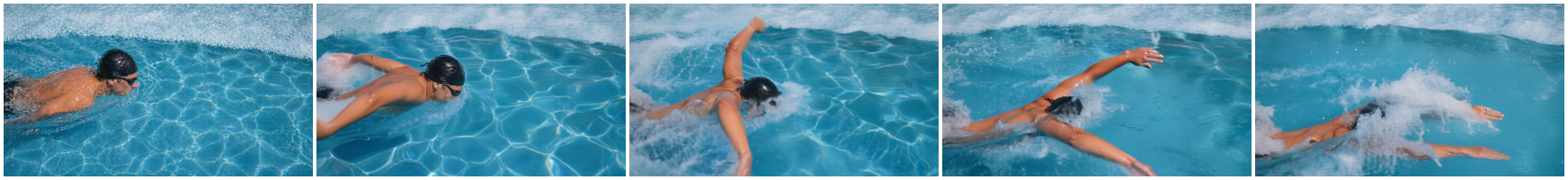}{Wan}

\caption{\textbf{I2V Motion Transfer comparison.}
Results generated from an input image and the text prompt
\textit{``A swimmer is swimming in the pool.''}
while preserving the transferred motion dynamics. The first frame is shown in the left column.
}
\label{fig:c4}
\end{figure*}

\begin{figure*}[t]
\centering

{\Large\bfseries Cut \& Drag Comparison\par}
\vspace{1mm}

\setlength{\rowheightI}{0.105\textheight}

\setlength{\imheightI}{0.75\rowheightI}

\setlength{\labelwidthI}{0.035\linewidth}

\newcommand{\resultrowI}[2]{%
  \noindent
  \begin{minipage}[c][\imheightI][c]{\labelwidthI}
    \centering
    \rotatebox{90}{\small\bfseries #2}
  \end{minipage}%
  \hfill
  \begin{minipage}[c][\imheightI][c]{0.95\linewidth}
    \includegraphics[width=\linewidth,height=\imheightI]{#1}
  \end{minipage}%
  \par\vspace{0pt}%
}

\resultrowI{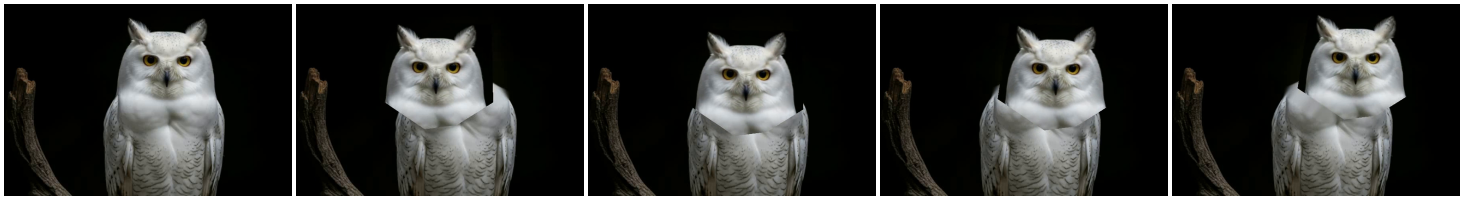}{Input}
\resultrowI{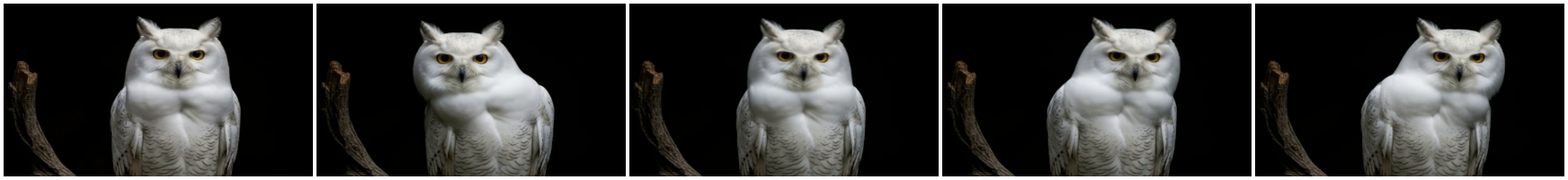}{Ours}
\resultrowI{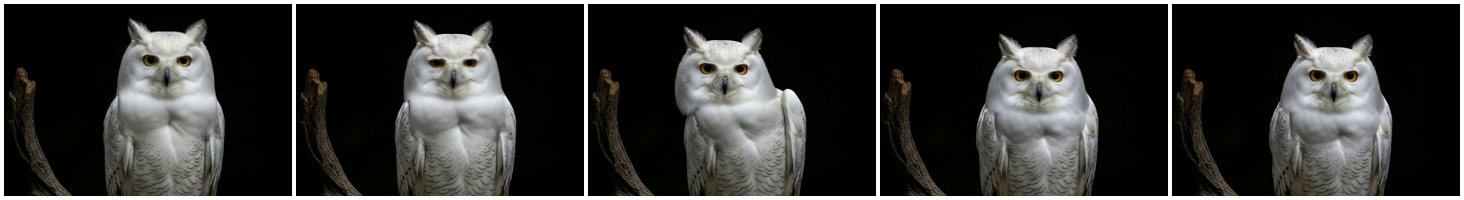}{ttm}
\resultrowI{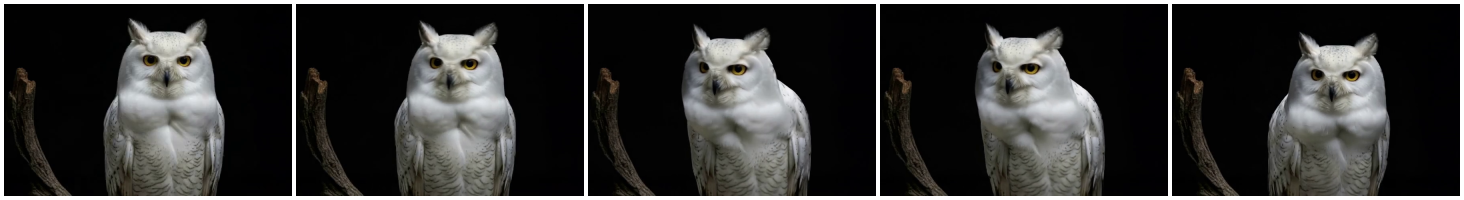}{GWTF}

\caption{\textbf{Cut \& Drag Comparison.}
Results generated from an input image and the text prompt
\textit{``A majestic snowy owl perches gracefully on a gnarled branch, its pristine white feathers adorned with delicate black speckles. The owl's piercing yellow eyes are wide and alert, scanning the surroundings with a sense of calm authority. As a gentle breeze rustles through the leaves, the owl remains poised, its sharp talons gripping the branch securely. The dark, blurred background accentuates the owl's striking presence, creating a serene yet powerful scene in the quiet of the night.''}
while preserving the transferred motion dynamics.
}
\label{fig:c5}
\end{figure*}

\begin{figure*}[t]
\centering

{\Large\bfseries T2V Motion Transfer Comparison\par}
\vspace{1mm}

\setlength{\rowheightT}{0.105\textheight}

\setlength{\imheightT}{0.75\rowheightT}

\setlength{\labelwidthT}{0.035\linewidth}

\newcommand{\resultrowT}[2]{%
  \noindent
  \begin{minipage}[c][\imheightT][c]{\labelwidthT}
    \centering
    \rotatebox{90}{\small\bfseries #2}
  \end{minipage}%
  \hfill
  \begin{minipage}[c][\imheightT][c]{0.95\linewidth}
    \includegraphics[width=\linewidth,height=\imheightT]{#1}
  \end{minipage}%
  \par\vspace{0pt}%
}

\resultrowT{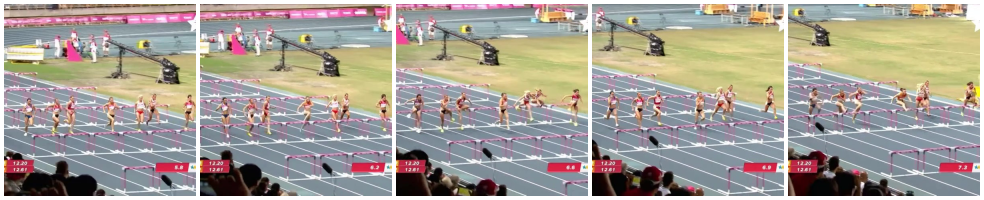}{Input}
\resultrowT{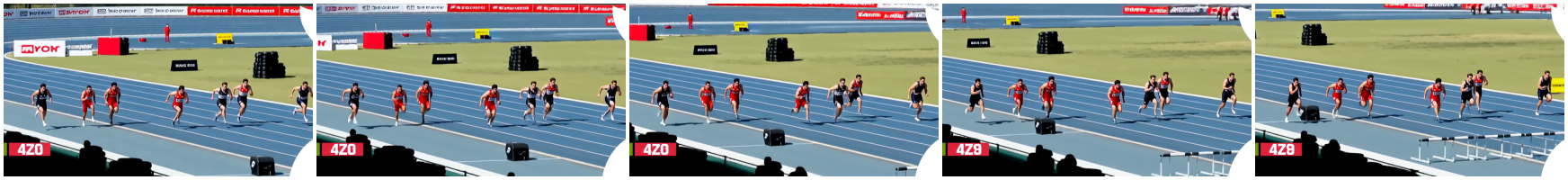}{Ours}
\resultrowT{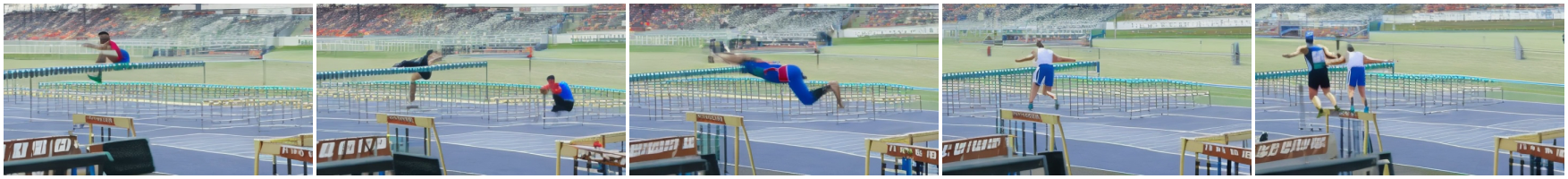}{DMT}
\resultrowT{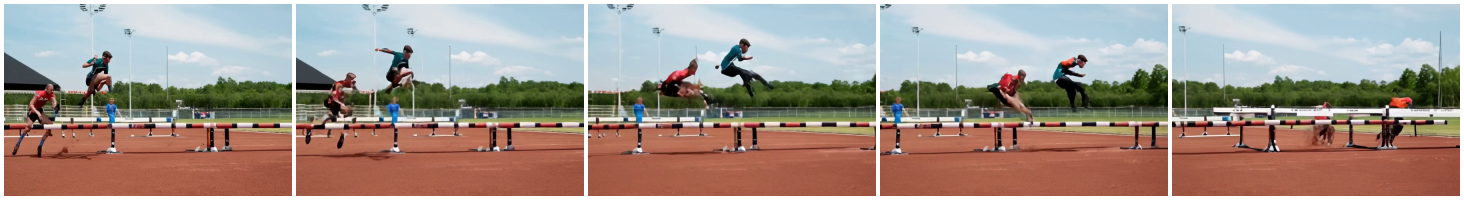}{DiTFlow}
\resultrowT{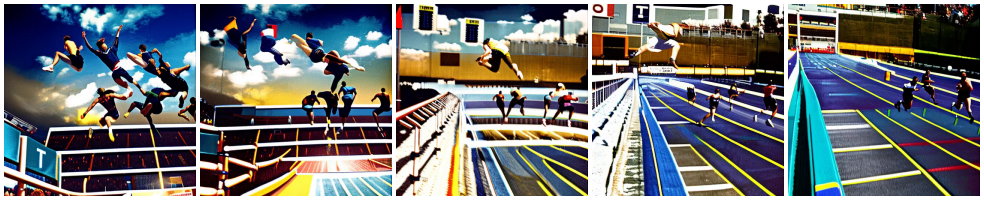}{MotionClone}
\resultrowT{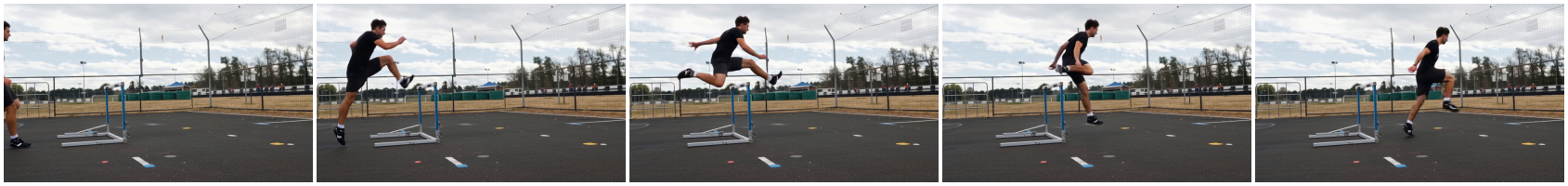}{Wan}

\caption{
\textbf{T2V Motion Transfer comparison.}
Qualitative comparison between different methods using the prompt
\textit{``Men jump over hurdles on a racetrack.''}
while preserving the motion dynamics from the input video.
}
\label{fig:c6}
\end{figure*}

\begin{figure*}[t]
\centering

{\Large\bfseries I2V Comparison\par}
\vspace{1mm}

\setlength{\rowheightI}{0.105\textheight}

\setlength{\imheightI}{0.75\rowheightI}

\setlength{\labelwidthI}{0.035\linewidth}

\newcommand{\resultrowI}[2]{%
  \noindent
  \begin{minipage}[c][\imheightI][c]{\labelwidthI}
    \centering
    \rotatebox{90}{\small\bfseries #2}
  \end{minipage}%
  \hfill
  \begin{minipage}[c][\imheightI][c]{0.95\linewidth}
    \includegraphics[width=\linewidth,height=\imheightI]{#1}
  \end{minipage}%
  \par\vspace{0pt}%
}

\resultrowI{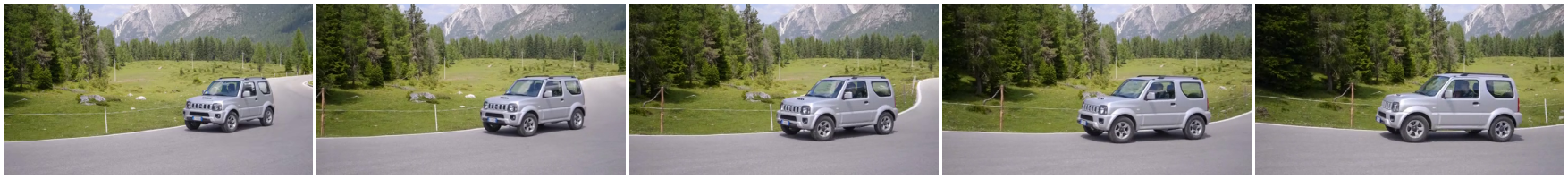}{Input}
\resultrowI{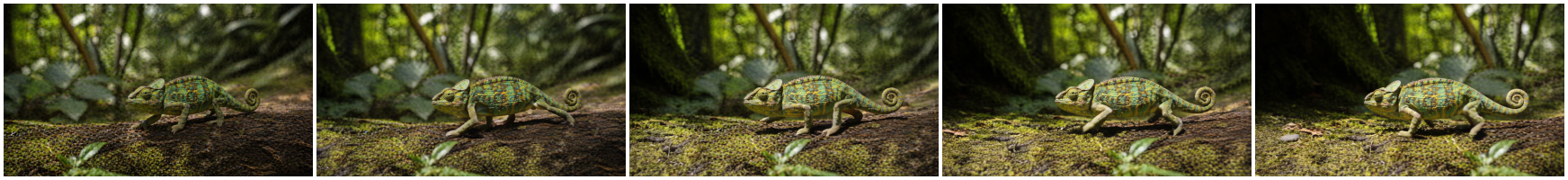}{Ours}
\resultrowI{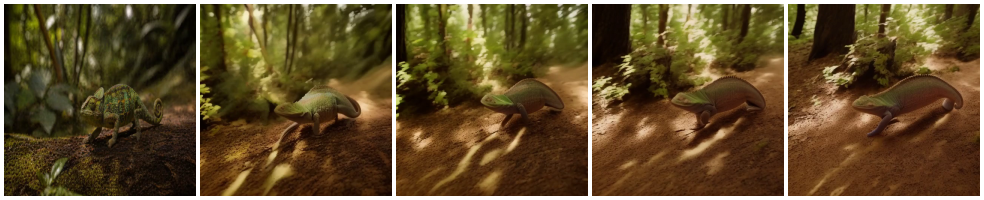}{MotionClone}
\resultrowI{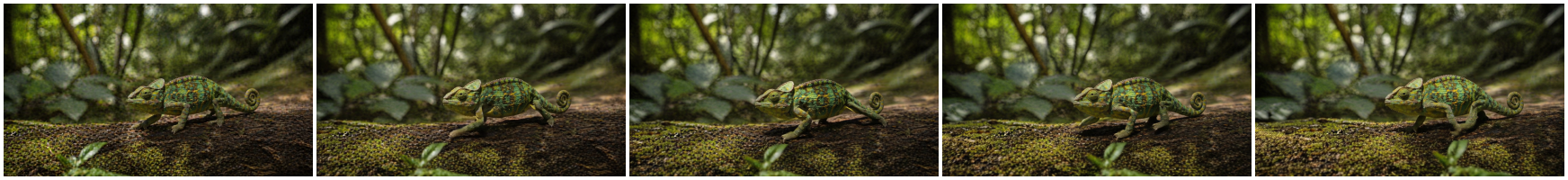}{Wan}

\caption{\textbf{I2V Motion Transfer comparison.}
Results generated from an input image and the text prompt
\textit{``A chameleon is walking in the forest.''}
while preserving the transferred motion dynamics. The first frame is shown in the left column.
}
\label{fig:c7}
\end{figure*}

\begin{figure*}[t]
\centering

{\Large\bfseries Cut \& Drag Comparison\par}
\vspace{1mm}

\setlength{\rowheightI}{0.105\textheight}

\setlength{\imheightI}{0.75\rowheightI}

\setlength{\labelwidthI}{0.035\linewidth}

\newcommand{\resultrowI}[2]{%
  \noindent
  \begin{minipage}[c][\imheightI][c]{\labelwidthI}
    \centering
    \rotatebox{90}{\small\bfseries #2}
  \end{minipage}%
  \hfill
  \begin{minipage}[c][\imheightI][c]{0.95\linewidth}
    \includegraphics[width=\linewidth,height=\imheightI]{#1}
  \end{minipage}%
  \par\vspace{0pt}%
}

\resultrowI{figures/main/CND/inputs/labobo_5.png}{Input}
\resultrowI{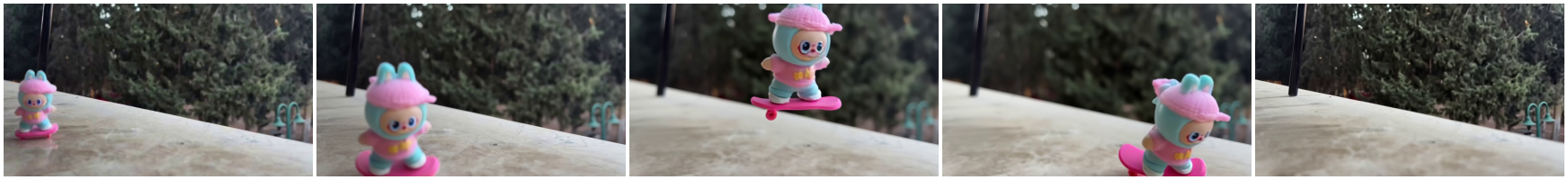}{Ours}
\resultrowI{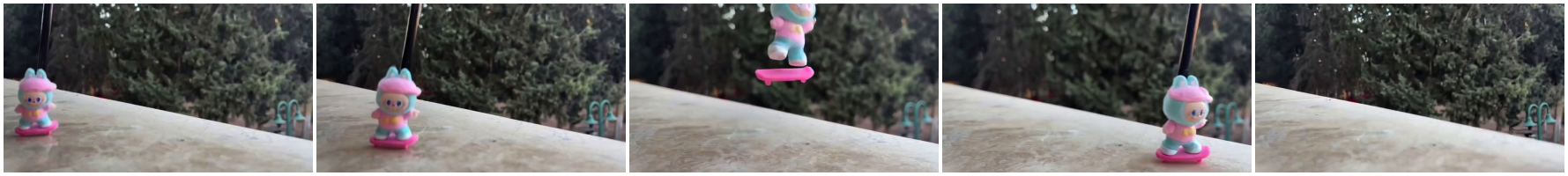}{ttm}
\resultrowI{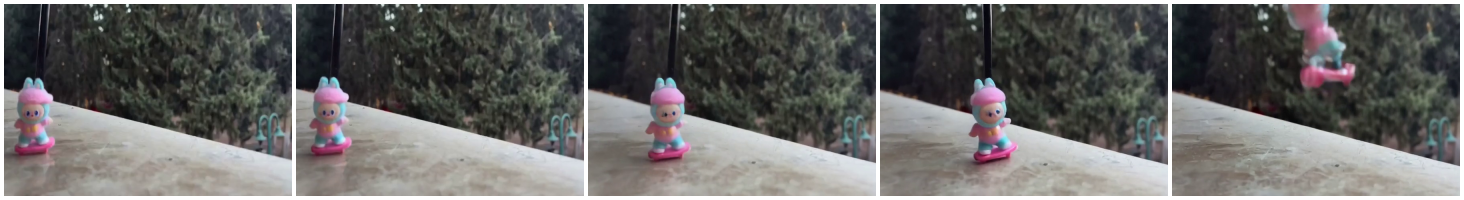}{GWTF}

\caption{\textbf{Cut \& Drag Comparison.}
Results generated from an input image and the text prompt
\textit`{`The toy is riding a miniature pink skateboard along a light-colored stone ledge. Against a blurred background of green trees. Midway through the scene, the skateboard jump, before it lands back on the ledge and continues its ride out of the frame.''}
while preserving the transferred motion dynamics.
}
\label{fig:c8}
\end{figure*}

We extend our comparisons presented in the main manuscript with additional samples. We present these comparisons in \cref{fig:c0,fig:c1,fig:c2,fig:c3,fig:c4,fig:c5,fig:c6,fig:c7,fig:c8}.

\subsection{$\gamma$ and $k$ Ablation}
\begin{figure*}[t]
    \centering
    \includegraphics[width=\textwidth]{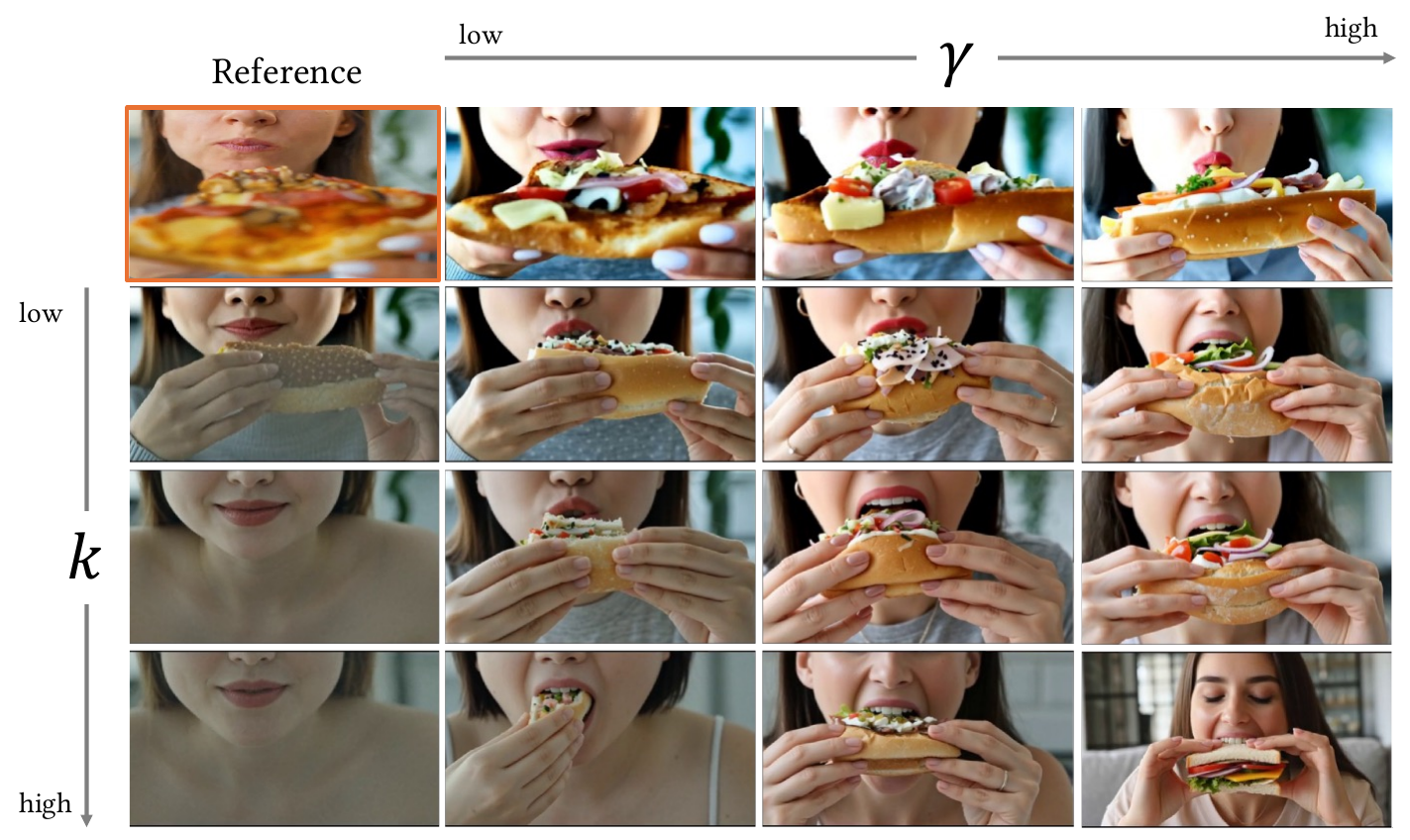}
    \caption{\textbf{$\gamma$ and k Ablation.} Demonstration of the effect of different $\gamma$ and $k$ combinations. The reference image is shown in the top-left corner. Prompt: \textit{``A woman eating a sandwich.''} }
    \label{fig:alb_1}

    \includegraphics[width=\textwidth]{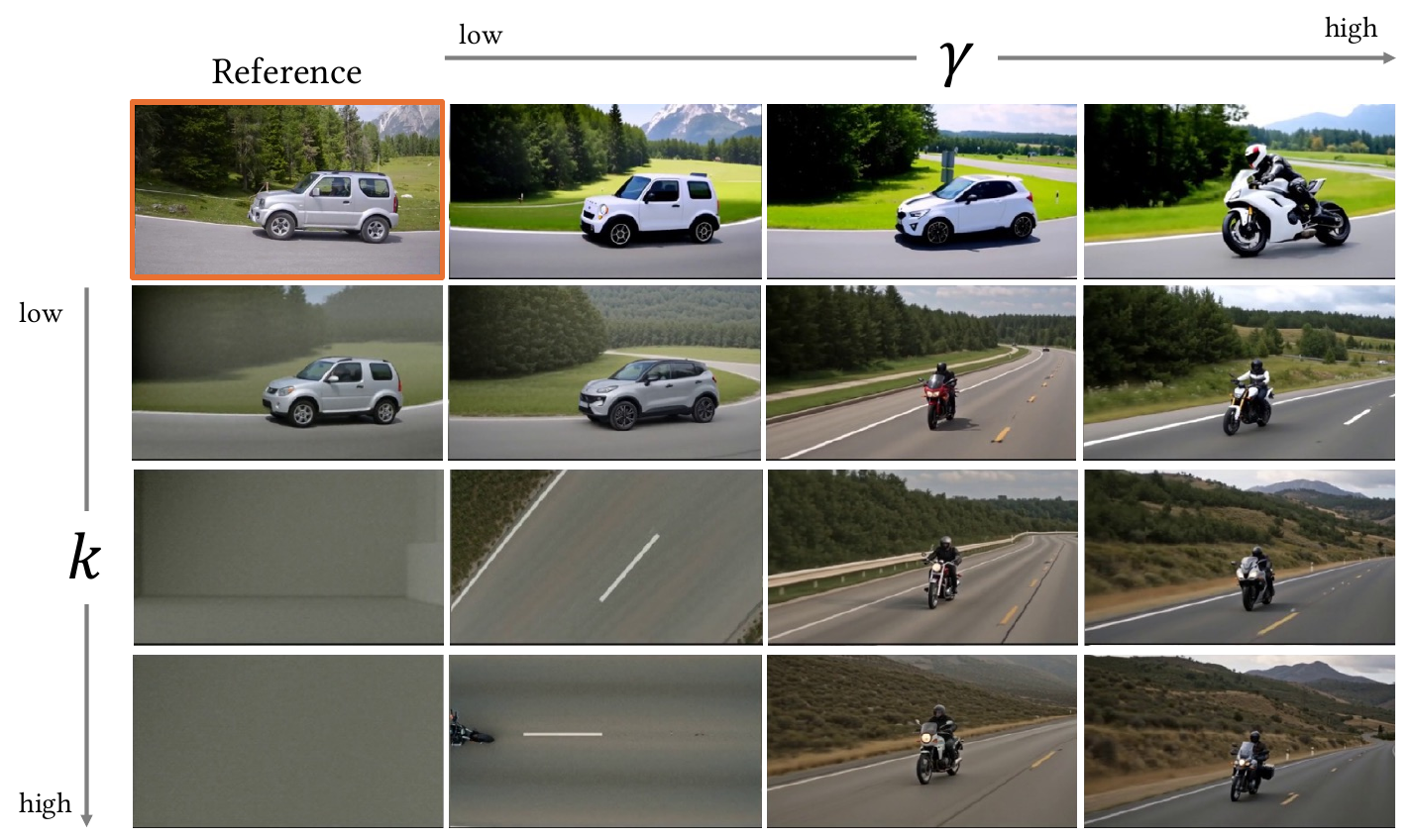}
    \caption{\textbf{$\gamma$ and k Ablation.} Demonstration of the effect of different $\gamma$ and $k$ combinations. The reference image is shown in the top-left corner. Prompt: \textit{``A man riding a motorcycle.''} }
    \label{fig:alb_2}
\end{figure*}

We illustrate the effect of varying the $\gamma$ and $k$ parameters in \cref{fig:alb_1,fig:alb_2}. As observed, different parameter combinations produce varying tradeoffs between visual fidelity and motion alignment.

\end{document}